\documentclass[10pt,twocolumn,letterpaper]{article} 
\usepackage[pagenumbers]{cvpr} 

\usepackage{graphicx}
\usepackage{amsmath}
\usepackage{amssymb}
\usepackage{booktabs}

\usepackage[dvipsnames]{xcolor}
\usepackage{colortbl}
\usepackage{footnote}
\usepackage{accents}
\usepackage{multirow}

\usepackage{BOONDOX-uprscr} 

\usepackage[linesnumbered,ruled,vlined]{algorithm2e}
\usepackage{listings}

\usepackage{standalone} 

\newcommand{\textcount}[1]{\texttt{#1}}
\newcommand{\textvec}[1]{\mathbf{#1}}
\newcommand{\textfunc}[1]{\mathscr{#1}}
\newcommand{\textset}[1]{\mathcal{#1}}
\newcommand{\textstat}[1]{\textsf{#1}} 
 
\newcommand{\textaggfunc}[1]{\textsc{#1}}
\newcommand{\textattack}[1]{\accentset{\bigtriangledown}{#1}}
\newcommand{\starhat}[1]{\accentset{\star}{#1}}

\newcommand{\paperref}[1]{\textit{\textcolor{PineGreen}{#1}}}

\newcommand{\without}{\textit{(w/o) }}

\definecolor{Tred}{rgb}{0.85, 0.3, 0.3}            
\definecolor{Torange}{rgb}{1.0, 0.65, 0.00}        
\definecolor{Tyellow}{rgb}{1.0, 0.95, 0.70}       
\definecolor{mbzblue}{rgb}{0.2431, 0.6745, 0.8745}
\definecolor{msugreen}{rgb}{0.0, 0.5372, 0.2040}

\makeatletter
\def\widebreve{\mathpalette\wide@breve}

\def\wide@breve#1#2{\sbox\z@{$#1#2$}%
     \mathop{\vbox{\m@th\ialign{##\crcr
\kern0.08em\brevefill#1{0.8\wd\z@}\crcr\noalign{\nointerlineskip}%
                    $\hss#1#2\hss$\crcr}}}\limits}
\def\brevefill#1#2{$\m@th\sbox\tw@{$#1($}%
  \hss\resizebox{#2}{\wd\tw@}{\rotatebox[origin=c]{90}{\upshape(}}\hss$}
\makeatletter

\definecolor{cvprblue}{rgb}{0.21,0.49,0.74}
\usepackage[pagebackref,breaklinks,colorlinks,allcolors=cvprblue]{hyperref}

\usepackage[capitalize]{cleveref}
\usepackage{comment}
\crefname{section}{Sec.}{Secs.}
\Crefname{section}{Section}{Sections}
\Crefname{table}{Table}{Tables}
\crefname{table}{Tab.}{Tabs.}

\def\paperID{12629} 
\def\confName{CVPR}
\def\confYear{2025}

\title{FedSECA: Sign Election and Coordinate-wise Aggregation of Gradients \\ for Byzantine Tolerant Federated Learning}

\author{
  Joseph Geo Benjamin$^{\dagger}$ \quad 
  Mothilal Asokan$^{*}$ \quad 
  Mohammad Yaqub$^{*}$ \quad 
  Karthik Nandakumar$^{* \dagger}$ \\
  \\
  $^{\dagger}$Michigan State University (MSU), East Lansing, MI, USA \\
  \color{msugreen}{\tt\small \{benja161, nandakum\}@msu.edu} \\
  $^{*}$Mohamed bin Zayed University of Artificial Intelligence (MBZUAI), Abu Dhabi, UAE \\
  \color{mbzblue}{\tt\small \{mothilal.asokan, mohammad.yaqub, karthik.nandakumar\}@mbzuai.ac.ae} \\
}

\begin{document}

\maketitle

\begin{abstract}
One of the most common defense strategies against Byzantine clients in federated learning (FL) is to employ a robust aggregator mechanism that makes the training more resilient. While many existing Byzantine robust aggregators provide theoretical convergence guarantees and are empirically effective against certain categories of attacks, we observe that certain high-strength attacks can subvert the robust aggregator and collapse the training. To overcome this limitation, we propose a method called FedSECA for robust \underline{s}ign \underline{e}lection and \underline{c}oordinate-wise \underline{a}ggregation of gradients in FL that is less susceptible to malicious updates by an omniscient attacker. The proposed method has two main components. The \textbf{Concordance Ratio Induced Sign Election} (CRISE) module determines the consensus direction (elected sign) for each individual parameter gradient through a weighted voting strategy. The client weights are assigned based on a novel metric called concordance ratio, which quantifies the degree of sign agreement between the client gradient updates. Based on the elected sign, a \textbf{Robust Coordinate-wise Aggregation} (RoCA) strategy is employed, where variance-reduced sparse gradients are aggregated only if they are in alignment with the corresponding elected sign. We compare our proposed FedSECA method against 10 robust aggregators under 7 Byzantine attacks on 3 datasets and architectures. The results show that existing robust aggregators fail for at least some attacks, while FedSECA exhibits better robustness. Code - {\small \hypersetup{urlcolor=magenta}{\url{{https://github.com/JosephGeoBenjamin/FedSECA-ByzantineTolerance}}} }

\end{abstract}



\section{Introduction}

Federated learning (FL) \cite{fed-mcmahan2017communication} enables collaborative training of a machine learning (ML) model by sharing statistical updates learned on data instead of sharing actual data with other clients or server, thus mitigating systemic privacy risks \cite{fed-kairouz2021advances}. FL has emerged as a vital tool for leveraging decentralized data, especially in applications with strict data storage and sharing regulations (e.g., GDPR). 



When entities collaboratively train ML models, they become a potential target for malicious actors. These actors may include third parties seeking to disrupt the system through sophisticated man-in-the-middle style attacks, which intercept and alter the transmitted updates. Alternatively, competing participating entities may also seek to undermine or sabotage the collaboration to gain a competitive advantage in the market. We refer to these clients that defect intentionally or unintentionally as \textit{Byzantines}.

Model poisoning is an extensively studied threat in FL, where Byzantines send in corrupt or tailored updates to derail the entire training process \cite{attk_jere2020taxonomy}. The impact of these attacks becomes more pronounced when the number of Byzantines is \textbf{near-majority} and the attacker is \textbf{omniscient} (with knowledge of data distribution, other honest updates and aggregation methods). Robust aggregation (\textaggfunc{RAggr}) is a fundamental defense strategy to make training resilient to Byzantine attacks. Though many \textaggfunc{RAggr} methods have been proposed for FL, fundamental issues exist when they are adapted for practical use.

\noindent\textbf{(1)} Middle-seeking \cite{defns-blanchard2017machine_krum,defns-yin2018byzantine_cwtm, defns-mhamdi2018hidden_bulyan} and outlier-suppression \cite{defns-alkhunaizi2022suppressing_copod} \textaggfunc{RAggr} methods by their design consider only a small subset of clients while suppressing/discarding the rest, using central tendency or mutual deviation of clients. 
These approaches do not accommodate the richness of data heterogeneity, a primary motive for collaboration, and become highly sub-optimal in the presence of large heterogeneity even without Byzantines. This is due to the underlying assumption that larger deviations are due to maliciousness when, in fact, they could be due to large heterogeneity. Furthermore, once the data distribution is known, attacks can be crafted to hide among the variance of honest clients~\cite{attk-hejwalkar2021minmax_minsum_dnc}, especially under the assumption of an omniscient attacker.

\begin{figure}[t]
\begin{center}
\includegraphics[width=\linewidth]{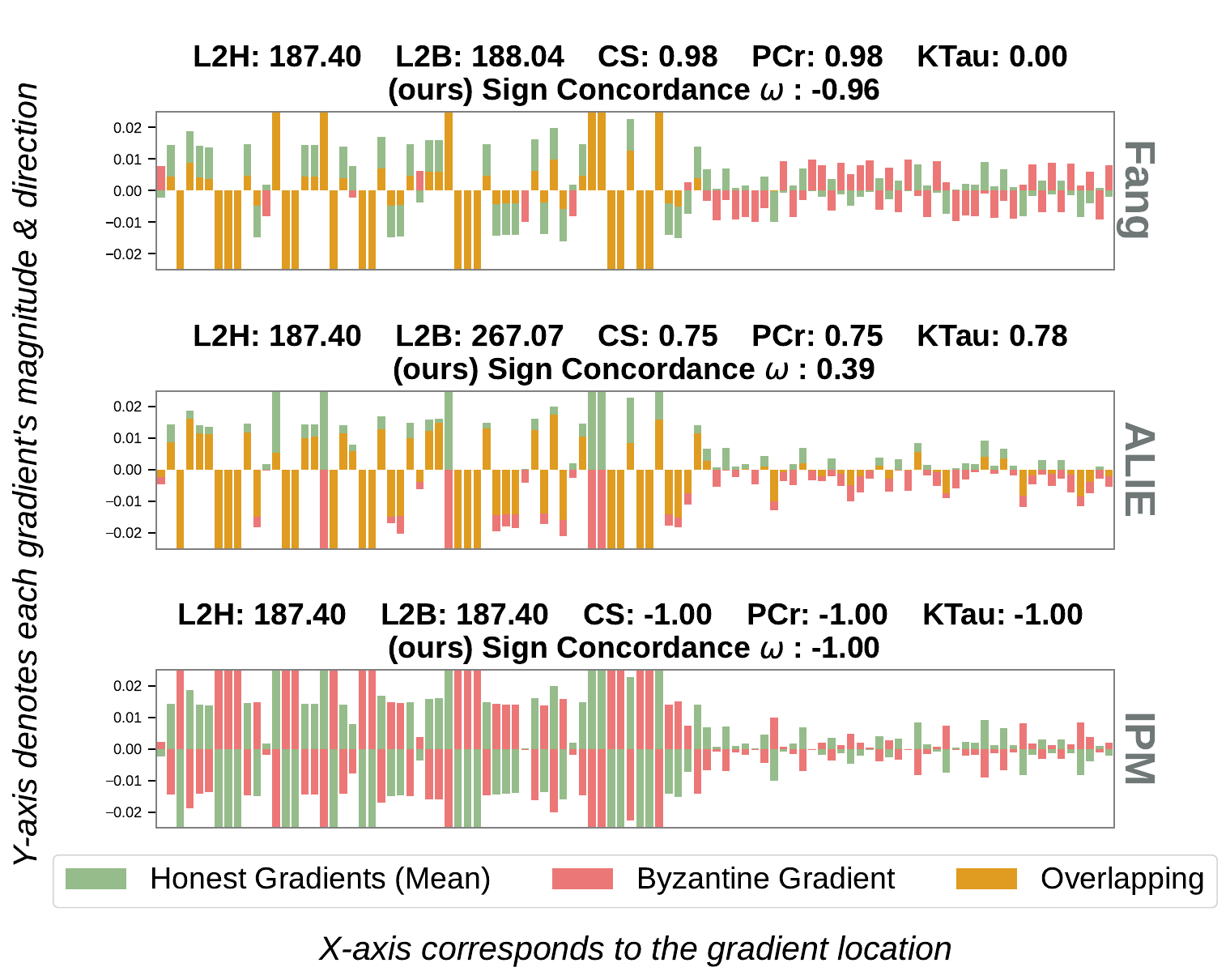}
\end{center}
   \caption{\textbf{\textit{Assessing similarity between Honest \& Byzantine gradients}}: In this figure, green represents the mean of honest gradients, red represents Byzantine gradients, and orange represents the overlap between the two. Clearly, metrics such as L2-norm are inadequate because attacks such as Fang and ALIE can result in similar L2-norms for honest (L2H) and Byzantine (L2B) gradients, though they do not align well directionally. Other metrics such as cosine similarity (CS), Pearson's correlation (PCr), and Kendall's rank correlation (KTau) also do not work well consistently across different attacks. In contrast, the sign concordance~($\omega$) metric proposed in this work gives a better indication of the degree of alignment between honest \& Byzantine gradients. Displayed gradients are from a subset of parameters in the ResNet18 model while federated learning on the CIFAR10 dataset.
   }
\label{fig:stat_of_attack}
\end{figure}

\noindent\textbf{(2)} Variance-reduction approaches like trimmed mean \cite{defns-yin2018byzantine_cwtm}, centered-clipping \cite{defns-karimireddy2021learning_clipping}, and bucketing \cite{defns-karimireddy2020byzantine_randBuck, defns-ozfatura2023byzantines_seqbuck,defns-allouah2023fixingbymixing} attempt to curtail the extreme malicious updates without excluding any clients. These methods include gradients from clients in the tail distributions, thus enabling access to richer gradients. Clipping only considers radius (L2-norm), making it oblivious to the angular/sign variance of update vectors (see \cref{fig:stat_of_attack}). Bucketing helps in controlling the variance before clipping \cite{defns-karimireddy2020byzantine_randBuck} and provides isolation among clients with angular variations \cite{defns-ozfatura2023byzantines_seqbuck}. Although the intensity of malicious gradients is reduced, the direction of the update is still perturbed, which can easily overpower the direction of the global update, leading to training collapse. This is especially problematic in cross-silo FL (with fewer clients) because each bucket may end up with a malicious vector corrupting the outputs of aggregation.

\noindent\textbf{(3)} Finally, robust aggregators often rely on certain hyperparameters (e.g., making initial judgments on the number of Byzantines). 
Some methods require prior knowledge of variance among updates or empirically estimate the variance for a given data distribution and FL setup. Practically, it is infeasible to rely on this knowledge because Byzantines can very well skew these estimates from the beginning.

In this work, we address training-time model poisoning attacks by omniscient attackers that significantly degrade global model performance in cross-silo FL settings. We specifically focus on developing an aggregation protocol \cite{defns-xie2023mixed_nash,defns-zhang2022flip_datapoisoning,defns-rajput2019detox_robust} that is robust to high-strength attacks, even when close to majority of participants are Byzantines. The only assumption we impose is that honest clients are a majority ($>50\%$), which is a common assumption across most robust aggregators. Our main contributions are:


\begin{itemize}    
    \item We propose a weighted voting strategy called CRISE to robustly determine the consensus direction (elected sign) for each parameter gradient in a cross-silo FL setup. The CRISE module leverages a novel metric called sign \textit{concordance ratio}, which quantifies the degree of directional alignment within a set of client gradient updates.
    
    \item We propose a robust coordinate-wise aggregation (RoCA) method that performs aggregation of variance-reduced sparse (VRS) gradients, where individual gradients are aggregated only if they align with the elected sign.

    \item By combining CRISE and RoCA, we obtain a robust aggregator called FedSECA for gradient aggregation in FL. We benchmark FedSECA against state-of-the-art defenses using well-known attacks and empirically demonstrate that FedSECA is resilient even under strong attacks. 
\end{itemize}


\begin{figure*}[t]
\begin{center}
\includegraphics[width=0.95\linewidth]{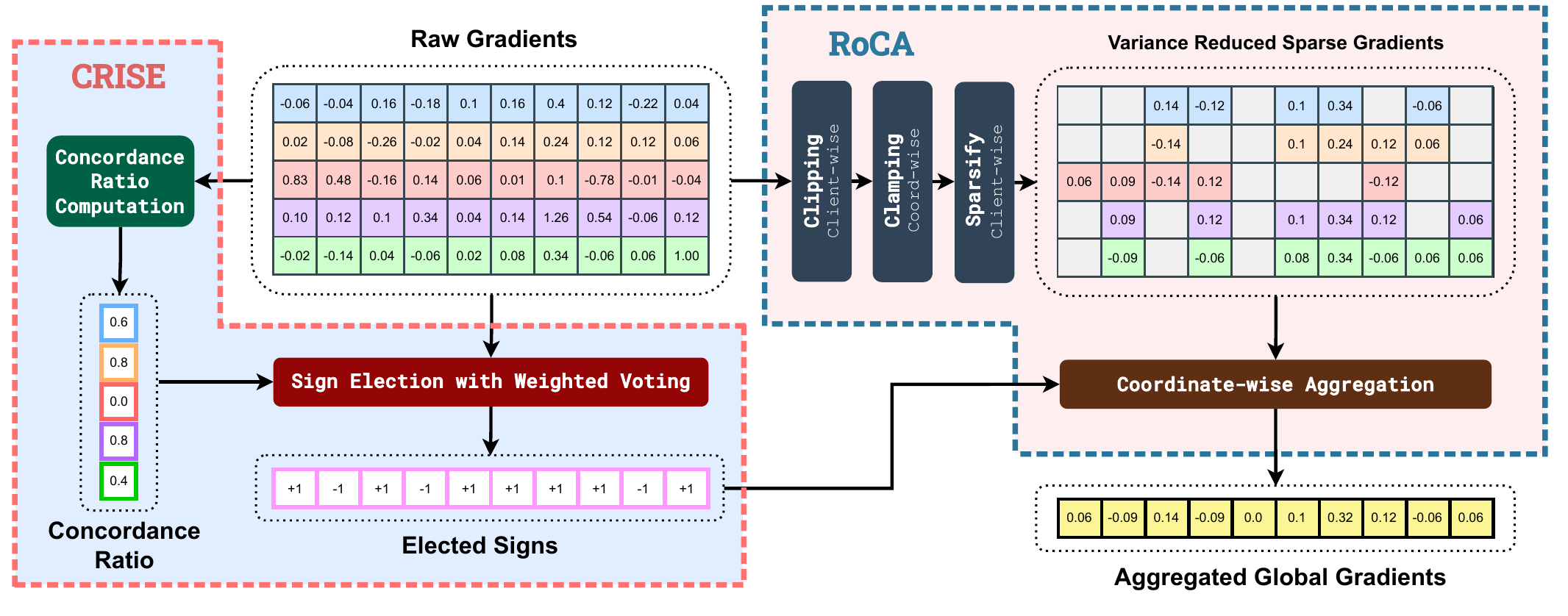}
\end{center}
   \caption{\textbf{\textit{Illustration of the proposed FedSECA approach for robust gradient aggregation in FL}}. The Concordance Ratio Induced Sign Election (CRISE) module determines the consensus direction (elected sign) for each parameter gradient through a weighted voting strategy. The clipping, clamping, and sparsification steps result in variance-reduced sparse gradients. The Robust Coordinate-wise Aggregation (RoCA) module aggregates individual gradients only when they are in alignment with the corresponding elected sign. Thus, FedSECA accommodates useful gradients from honest heterogeneous clients while suppressing the impact of gradient updates from Byzantine clients.}
\label{fig:fedrise_method}
\end{figure*}

\section{Related Work}

\noindent \textbf{Model Poisoning Attacks in FL}: 
Failure modes of typical deep learning systems such as adversarial \cite{lit-nair2023robust_adversarial}, backdoor \cite{lit-nguyen2024backdoor}, and data poisoning \cite{lit-tolpegin2020datapoison} attacks are also applicable to FL systems. Additionally, the collaborative nature of FL exposes these systems to new vulnerabilities like model poisoning attacks, which are caused by the corruption of updates shared by the clients during aggregation. 
One of the easiest attacks is random perturbation \cite{defns-blanchard2017machine_krum}, which causes training to diverge, but it is easier to detect. \textbf{Scaling} attacks resize gradients along any given direction, causing gradient explosion or divergence. In the \textbf{LabelFlip} attack, models are trained on constantly flipped labels \cite{attk-biggio2012poisoning_labelflip}. Examples of stealthy attacks designed to sabotage the training without being identified as malicious include \textbf{IPM} \cite{attk-xie2020fall}~-~pushes gradients exactly opposite to the average update direction, and Fang \etal \cite{attk-fang2020local} (henceforth, referred to as \textbf{Fang})~-~crafts parameter direction such that gradients below certain threshold are made opposite to the optimization objective with high intensity. Attacks that are specifically targeted against robust aggregators (especially middle-seeking approaches) include \textbf{ALIE} \cite{attk-baruch2019little}~-~perturbs the gradients slightly in different directions each time within bounds of honest updates, and \textbf{Mimic} \cite{defns-karimireddy2020byzantine_randBuck}~-~alters the imbalance perceived by aggregators by duplicating the updates of outlier clients. 
Moreover, optimization-based attacks like \textbf{MinMax}\cite{attk-hejwalkar2021minmax_minsum_dnc} or PoisonedFL\cite{attk-xie2024poisonedfl} dynamically compute attack parameters at each step of aggregation to increase effectiveness.

\noindent \textbf{Robust Aggregators}: To defend against model poisoning attacks in FL, numerous robust aggregation (\textaggfunc{RAggr}) methods have been proposed and they can be broadly categorized into five groups. 
\textbf{\textit{ (a) Middle-seeking}} methods like Krum \cite{defns-blanchard2017machine_krum} and geometric median-based aggregation (RFA)~\cite{defns-pillutla2022rfa_geomed} use centrality measures like median of clients updates to select a central update. They assume that the selected update is less likely to be malicious and will be representative of the entire population. Coordinate-wise trimmed mean and median (CWTM) \cite{defns-yin2018byzantine_cwtm} uses the central tendency of individual parameter gradients (instead of treating the entire gradient vector as a single entity) to ignore extreme values. Bulyan\cite{defns-mhamdi2018hidden_bulyan} is a combination of Krum and CWTM. 
Huber Loss minimization (HuberLoss) \cite{defns-zhao2024huberloss} method handles both consistency and robustness of models by utilizing both geometric median and L2-distance measures. 
While middle-seeking methods are effective against random update faults and simple attacks because they discard the gradients of most clients, omniscient attackers can collude to craft malicious updates that would be selected by the median methods and thereby slowly diverge the training. 
\textbf{\textit{(b)~Centered-Clipping~\&~Bucketing}} methods \cite{defns-karimireddy2021learning_clipping, defns-karimireddy2020byzantine_randBuck, defns-ozfatura2023byzantines_seqbuck, defns-allouah2023fixingbymixing} do not remove or discard clients, but limit the variation of the individual client updates to the extent that the harmful updates are neutralized among other clients. This approach typically ensures convergence even under heterogeneity, as they do not harshly discard clients as in the middle-seeking strategy. The Bucketing strategy splits clients into buckets based on some heuristics, thus enabling the mixing of updates and ensuring that the likelihood of Byzantines forming the majority is small. 
\textbf{\textit{(c) Clustering \& Outlier detection}} methods like SignGuard \cite{defns-xu2022byzantine_signguard} and other explicit clustering approaches \cite{defns-sattler2020byzantine_cosineclustering, defns-li2023experimental_clippedclustering} use some form of similarity assessment like L2-norm or cosine similarity to cluster the client updates. Then they choose the dominant or robust group for aggregation.
Methods like COPOD-DOS \cite{defns-alkhunaizi2022suppressing_copod} or FLDetector \cite{defns-zhang2022fldetector}
detect and discard the outliers in the updates using statistical measures. RFFL \cite{defns-xu2020reputation_RFFL} uses reputation computed based on cosine similarity to assess the contribution and reliability of each client's gradients.
\textbf{\textit{(d) Sign Voting}}:
SignSGD-MV \cite{defns-bernstein2018signsgd} proposed quantized (0~-~1) gradient aggregation for communication efficiency with a voting mechanism for Byzantine tolerance. Further \cite{defns-sohn2020election_coding_signsgd} suggested an improved election coding strategy for improving the robustness of SignSGD. A similar formulation has been used in model merging  \cite{merge-yadav2023resolving_ties}, where models trained on different tasks are merged after resolving sign conflicts. 
\textbf{\textit{(e)~Multi-Aggregator Frameworks}}:
Detox \cite{defns-rajput2019detox_robust} proposes a hierarchical framework for multistage polling among different groups, where aggregators are used at different levels to prevent targeted attacks.
RobustTailor \cite{defns-xie2023mixed_nash} formulates robust FL as a game between attacker and server, wherein the server can choose any \textaggfunc{RAggr} from a given set.

While many robust aggregators provide theoretical convergence guarantees and are effective against some attacks, they are still inadequate against certain high-strength attacks. Hence, there is a need for stronger \textaggfunc{RAggr} methods that can handle large heterogeneity and complex models. Concurrent work Xu.\etal \cite{defns-xu2024LASAByzantine} uses sparsified layerwise aggregation with positive direction purity. Our findings Table.\ref{tab:fedrise_components} show that sparsification alone cannot prevent direction-aligned scaling attacks, emphasizing the importance of the VRS Gradients. Our method does not make assumptions like the predominance of positive gradients nor need any architectural/layer information for aggregation.


\section{Problem Setup} \label{sec:problem-definitions}

\noindent \paragraph{Preliminaries:} Let $\textset{K} = \{1,2,\cdots,\textcount{K}\}$ represent the set of all clients involved in the FL setup, where $\textcount{K}$ is the number of clients. Let $\textset{D}_k = \{(\mathbf{x}_i^k,y_i^k)\}_{i=1}^{\textcount{N}_k}$ represent the local dataset with client $k$, where $\textcount{N}_k$ is the number of samples available at the $k^{th}$ client, $k \in \textset{K}$. Note that $\mathbf{x} \in \mathcal{X}$ and $y \in \mathcal{Y}$, where $\mathcal{X}$ and $\mathcal{Y}$ denote the input and output spaces of a machine learning (ML) model $\mathcal{M}_\textvec{w}:\mathcal{X}\rightarrow\mathcal{Y}$. Here, $\mathcal{M}$ denotes the model architecture, $\textvec{w} \in \mathbb{R}^\textcount{D}$ denotes the model parameters (i.e., weights), and $\textcount{D}$ is the number of parameters. Let $\textfunc{L}(\mathbf{x},y;\textvec{w})$ be the per-sample loss and $\textit{L}_k(\textvec{w}) = \frac{1}{\textcount{N}_k} \sum_{i=1}^{\textcount{N}_k} \textfunc{L}(\mathbf{x}_i^k,y_i^k;\textvec{w})$ be an estimate of the expected loss at client $k$. FL training is a distributed optimization problem \cite{fed-mcmahan2017communication}, where the goal is to learn the global model parameters $\textvec{w}$ such that:

\begin{equation}
    \min_{\textvec{w}} ~ \textit{L}(\textvec{w}) := \sum_{k\in\textset{K}} \alpha_k \textit{L}_k(\textvec{w}), \nonumber
\label{eq:fed}
\end{equation}

\noindent where $\alpha_k = \frac{\textcount{N}_k}{\sum_{j=1}^\textcount{K} {\textcount{N}_j}} \ \ \text{or}\ \ \frac{1}{\textcount{K}}$. Let $\nabla \textit{L}$ denote the true gradient of the above loss function $\textit{L}$ w.r.t. $\textvec{w}$.

In each round $t$ of FL training, the server broadcasts the previous model parameters $\textvec{w}^{(t-1)}$. Each client computes either the local model parameters $\textvec{w}_k^{(t)}$ or local gradients $\textvec{g}_k^{(t)}$. These local updates are aggregated by the server through an aggregation function $\textaggfunc{Aggr}(\cdot)$ to obtain the current model parameters $\textvec{w}^{(t)}$. For example, simple FedAvg aggregation function can be represented as follows:

\begin{equation}
   \textvec{w}^{(t)} \;= \textaggfunc{FedAvg} ( \{\textvec{w}^{(t)}_k : k \in \textset{K}\} )
   \;:= \sum_{k\in\textset{K}} \alpha_k \textvec{w}^{(t)}_k.
\end{equation}

\noindent Note that $t \in [1,\textcount{T}]$, where $\textcount{T}$ is the number of communication rounds and the model parameters $\textvec{w}^{(0)}$ for the first round are initialized randomly by the server.

\noindent\paragraph{Attacker Goal and Capability:}

The goal of Byzantine clients or malicious actors is to disrupt the collaborative training process by corrupting the updates, ultimately causing a denial-of-service attack on the ML task.
This can be achieved by pursuing a direction opposite to the true gradients $\nabla \textit{L}$ of optimization \cite{attk-xie2020fall} or scaling updates to cause exploding gradients. All the attackers are assumed to be \textit{\textbf{omniscient}} \ie, have full knowledge of the defensive aggregation in the server and access to all honest gradients computed by clients locally. Attackers also have local data distribution corresponding to the task at hand (for label-flipping attacks). The attacker can seduce honest clients by crafting updates closer to a specific set of honest clients, causing the aggregator to give more emphasis to the attacker, thereby excluding other honest clients. Let $\textset{B} \subset \textset{K}$ represent a set of Byzantines (malicious clients) and $\textset{H} \subset \textset{K}$ represent a set of honest clients. In this work, we assume that $\textset{B}$ and $\textset{H}$ form a partition of $\textset{K}$, i.e., $\textset{B}\cup\textset{H} = \textset{K}$ and $\textset{B}\cap\textset{H} = \emptyset$. Let $\textcount{B} = |\textset{B}|$ be the total number of Byzantine clients and $\textcount{H}=|\textset{H}| = (\textcount{K} - \textcount{B})$ be total number of honest clients. We assume that honest clients have a majority, i.e., $\textcount{H} > \textcount{B}$.

\noindent\paragraph{Robust Aggregation Problem Statement:}

In this work, we assume that the clients share local gradients $\textvec{g}_k^{(t)}$ in each round $t$ and the server aggregates them to obtain the global gradient $\textvec{g}^{(t)}$. The global model is then updated as $\textvec{w}^{(t)} = \textvec{w}^{(t-1)} - \eta \textvec{g}^{(t)}$, where $\eta$ is the global learning rate. Henceforth, we drop the index $t$ for simplicity. The baseline aggregation function\footnote{Note that the self-reported $\textcount{N}_k$ values are unreliable when some of the clients are Byzantines. Hence, $\alpha_k$ is set to $\frac{1}{K}$ during aggregation.} employed by the server is defined as:

\begin{equation}
\textvec{g} = \textaggfunc{Aggr}( \{\textvec{g}_k : k \in \textset{K}\}) := 
\frac{1}{\textcount{K}} \sum_{k \in \textset{K}} \textvec{g}_k.
\label{eq:baselineaggr-output}
\end{equation}


Clearly, the above baseline aggregator is inadequate in the presence of Byzantine clients. Hence, the objective of a \textit{robust aggregator} \textaggfunc{RAggr} is to return gradients $\widetilde{\textvec{g}}$ closer to the aggregation of honest clients $\textset{H}$, even in the presence of Byzantines $\textset{B}$. Thus, \textaggfunc{RAggr} is expected to filter out any harmful updates shared by Byzantine clients.

\begin{equation}
\widetilde{\textvec{g}} = \textaggfunc{RAggr}(\{\textvec{g}_k : k \in \textset{K}\}) \approx 
\frac{1}{\textcount{H}} \sum_{h \in \textset{H}} \textvec{g}_h.
\label{eq:robustaggr-output}
\end{equation}

In practice, designing a robust aggregator as described in Eq.~\ref{eq:robustaggr-output} is challenging because the aggregator is oblivious to the nature of the attack, the number of Byzantines, any collusion between Byzantines, or the expected true gradients.


\section{Designing a Robust Aggregator}

The gradients transmitted by the clients in a FL setup often have large variances among themselves, especially when the client data distributions are heterogeneous (non-iid). This large variance makes it easy for the attacker to hide the malicious updates. To mitigate this issue, we design a robust aggregator that consists of two key components: CRISE - Concordance Ratio Induced Sign Election and RoCA - Robust Coordinate-wise Aggregation. A block diagram summarizing our method is shown in \cref{fig:fedrise_method}.


\subsection{Concordance Ratio Induced Sign Election} \label{sec:crise_method}

Attacks that target the optimization direction can perturb mildly yet change the direction severely \cite{attk-baruch2019little}. One simple defense is to consider the gradients of all parameters as a single vector $\textvec{g} \in \mathbb{R}^{\textcount{D}}$ and find cosine similarity between gradient pairs across all clients \cite{defns-xu2020reputation_RFFL,defns-sattler2020byzantine_cosineclustering}. However, weights of over-parameterized deep neural network models typically follow a long-tailed distribution with a lot of parameters being negligible. Hence, attackers can falsify directional alignment by just aligning a few salient (high magnitude) parameters and corrupting the rest. Also, there are a large number of noisy gradients because of trivial/useless updates to parameters that do not contribute to learning. Hence, there is often no clear consensus among gradients, making identification of malicious clients difficult. 

To overcome the above problems, we propose a novel approach to \textit{determine a consensus direction (elected sign) for the gradient of each parameter in the global model}. Our approach is inspired by recent advances in the model merging literature such as TIES-Merging \cite{merge-yadav2023resolving_ties}, where the elected sign for each individual model parameter is first determined, and only the parameter values of those models that agree with the elected sign are aggregated. However, the presence of malicious clients makes sign election a harder problem in Byzantine-robust FL compared to benign model merging. Hence, we first propose a new metric called \textit{concordance ratio} to rate the client gradient updates and then perform sign election through weighted voting, where client weights are assigned based on their concordance ratio. 

\paragraph{Concordance Ratio:} For any two vectors $\textvec{g}_1,\textvec{g}_2 \in \mathbb{R}^{\textcount{D}}$, the sign concordance $\omega$ between them is defined as:

\begin{equation}
    \omega(\textvec{g}_1,\textvec{g}_2) = \frac{1}{\textcount{D}} \sum_{j=1}^{\textcount{D}} \textstat{sgn}(g_1^j)\cdot\textstat{sgn}(g_2^j), 
    \label{eq:signconcordance}
\end{equation}

\noindent where $\textvec{g}_1 = [g_1^1,g_1^2,\cdots,g_1^{\textcount{D}}]$, $\textvec{g}_2 = [g_2^1,g_2^2,\cdots,g_2^{\textcount{D}}]$, and $\textstat{sgn}$ is the signum function, i.e., any real $a = |a|\textstat{sgn}(a)$. Note that $\omega \in [-1,1]$ and sign concordance will be positive if the number of coordinate-wise sign agreements (concordances) between two vectors is more than the discordances.  

In the context of FL, given a set of gradient updates $\{\textvec{g}_k : k \in \textset{K}\}$ from $\textcount{K}$ clients, the sign \textit{concordance ratio} $\rho_k$ of client $k$ is defined as:

\begin{equation}
    \rho_k = \textstat{max} \Biggl[0, \; \frac{1}{\textcount{K}} \sum_{\ell \in \textset{K}} \textstat{sgn}\left(\omega(\textvec{g}_k,\textvec{g}_{\ell})\right)\Biggr].  
\end{equation}

\noindent Note that $\rho_k \in [0,1]$ and the minimum value of $\rho_k$ occurs when the gradient update of client $k$ is not in positive sign concordance with the gradient updates from a majority of other clients. In contrast, higher values of $\rho_k$ indicate good directional agreement between the gradient updates of client $k$ and the majority of other clients. Furthermore, under the honest majority assumption, honest clients can be expected to have higher $\rho_k$ and malicious clients that attempt to change the optimization direction will have lower $\rho_k$. 

\paragraph{Sign Election:} The main objective of this step is to determine the elected sign $\textvec{s} = [s^1,s^2,\cdots,s^D] \in \{-1,0,1\}^{\textcount{D}}$ for the gradient of each parameter in the model. The sign voting component is inspired by Election-Coding\cite{defns-sohn2020election_coding_signsgd} and TIES-Merging\cite{merge-yadav2023resolving_ties}. In Election-Coding, the final update is determined by election using quantized gradients, and it is mathematically similar to the bucketing strategy. But our problem addresses real-valued gradients with richer information. In TIES-Merging, the magnitude of the parameter (after sparsification/trimming) determines the voting weight. Since Byzantines can send large gradient values for individual parameters to skew the voting, we propose to employ the concordance ratio as the voting weight so that clients with a lower concordance ratio will contribute less during the voting. Hence, we refer to the proposed sign election technique as Concordance Ratio Induced Sign Election (CRISE), which can be expressed as:

\begin{equation}
    s^j = \textstat{sgn}\left(\sum_{k\in\textset{K}}\rho_k\cdot\textstat{sgn}(\starhat{g}_k^j)\right).
    \label{eqn:CRISE}
\end{equation}
Here, $\textvec{\starhat{g}}_k = [\starhat{g}_k^1,\starhat{g}_k^2,\cdots,\starhat{g}_k^{\textcount{D}}]$ is the final gradient of client $k$ that is passed to the aggregation. It is derived from the VRS gradient computation, as described in the next section.

\subsection{Coordinate-wise Robust Aggregation}

After electing the sign of each parameter gradient $\textvec{s}$, only the client gradients that align with the corresponding elected sign are selected for coordinate-wise aggregation, while the rest are ignored. This ensures that conflicting gradients do not derail or slow the convergence.

\begin{equation}
\begin{gathered}
    \delta_k^j = I(s^j \cdot \starhat{g}_k^j > 0), \\
    \tilde{g}^j = \frac{\sum_{k \in \textset{K}} \delta_k^j \cdot \starhat{g}_k^j}{\sum_{k \in \textset{K}} \delta_k^j}, 
    \end{gathered}
\label{eq:fedrise-alignaggr}
\end{equation}

\noindent where $I(z)$ is the indicator function which is equal to $1$ if $z$ is true, and $0$ otherwise. Here, $j = 1,2,\cdots,\textcount{D}$ and $\widetilde{\textvec{g}} = [\tilde{g}^1,\tilde{g}^2,\cdots,\tilde{g}^{\textcount{D}}]$ is the global gradient after aggregation, which is meant to be shared back with clients.

\paragraph{Variance Reduction \& Gradient Sparsification}: While the coordinate-wise aggregation approach is designed to defend against attackers attempting to alter the update direction, it is still vulnerable to scaling attacks. Scaling attacks by Byzantine clients can be mitigated by clipping the L2-norm of the client updates within certain bounds. However, attackers can still craft gradients that satisfy the L2-norm criteria (shown in Fig.\ref{fig:stat_of_attack}) but sabotage the training. This can be achieved by scaling only a subset of smaller gradient values, while still obeying the norm bounds \cite{attk-fang2020local}, thereby causing instability in training. To defend against such attacks, we propose combining both clipping and clamping, followed by a sparsification step to obtain variance-reduced sparse (VRS) gradients before aggregation. 

In order to restrict the variance of the updates, we first clip the gradient vector. Unlike centered-clipping \cite{defns-karimireddy2021learning_clipping}, we do not make any assumption about the L2-norm (denoted by $\|\cdot\|_2$) bounds of the gradient vector. Instead, we use the median of all clients as a radius and clip the gradients as:
 
\begin{equation}
    \begin{aligned}
    \widehat{\textvec{g}}_k \ =&\ \textstat{clip}(\textvec{g}_k)
    \ :=\ \textvec{g}_k \cdot \left(\textstat{min}\Bigl[1, \frac{\tau}{\|\textvec{g}_k \|_2} \Bigr]\right), \\
    \text{where}\ &\tau =\ \textstat{median}( \{ \|\textvec{g}_k\|_2 : k \in \textset{K}\}).
 \end{aligned}
\end{equation}

While clipping reduces the overall scale of the gradient vector, it is also crucial to consider the scales of each individual parameter gradient. Unlike the Coordinate-wise trimming approach \cite{defns-yin2018byzantine_cwtm} where extreme gradient values along each coordinate are ignored, we do not ignore any gradients. Instead, we use the median of the gradient magnitude at each coordinate to clamp the magnitudes as:

\begin{equation}
    \begin{aligned}
    \overline{\textvec{g}}_k \ = &\ \textstat{clamp}(\widehat{\textvec{g}}_k) 
 := [\bar{g}_k^1,\bar{g}_k^2,\cdots,\bar{g}_k^{\textcount{D}}] \\
    &\bar{g}_k^j = \textstat{sgn}(\Hat{g}_k^j) \cdot \left(\textstat{min} \Bigl[ \mu^j , |\Hat{g}_k^j| \Bigr]\right),\\
    \text{where}\ &\mu^j =\ \textstat{median}(\{|\Hat{g}_k^j| : k \in \textset{K} \}).
\end{aligned}
\end{equation}

Next, to remove the noisy and trivial updates that might not be required for convergence, we select only the significant gradients. Such gradient sparsification has been studied for compressing gradients communicated in distributed paradigms with convergence guarantees\cite{merge-lin2017deepcompression}. These methods tend to sparsify $99\% - 99.9\%$ of gradients and with very marginal impact on convergence, but require error accumulation and momentum masking strategies to ensure that all gradients can be accounted for training to prevent any divergence. However, in the presence of Byzantine clients, adding such an error correction mechanism is detrimental as the error term would also include a poisoned term, which would accumulate and become dominant over time and corrupt the training.  Also, error accumulation need not be explicitly handled in FL, as each client can take multiple local steps sufficient to recover sparsification. The concept of sparsifying the weak gradients is also used in a single-step model merging \cite{merge-yadav2023resolving_ties} for eliminating the introduction of noise to existing weights. In this work, the server filters out a fraction $\gamma$ of gradient values and retains only the top $(1-\gamma)$ fraction of high magnitude gradients. 

\begin{equation}
    \begin{aligned}
    \textvec{\ddot{g}}_k \ = &\ \textstat{sparsify}_{\gamma}(\textvec{g}_k,\overline{\textvec{g}}_k) 
 := [\ddot{g}_k^1,\ddot{g}_k^2,\cdots,\ddot{g}_k^{\textcount{D}}] \\
    &\ddot{g}_k^j = \bar{g}_k^j \cdot \left(I(|g_k^j| > \lambda_k)\right),\\
    \text{where}\ &\lambda_k =\ \gamma\text{-}\textstat{quantile}(\{|g_k^1|,|g_k^2|,\cdots,|g_k^{\textcount{D}}|\}).
\end{aligned}
\end{equation}

Clipping, clamping, and sparsification operations are applied sequentially in that order to obtain VRS gradients. Note that while $\lambda_k$ and coordinates $j$ to sparsify are determined based on raw gradients $\textvec{g}_k$, sparsification is applied to $\overline{\textvec{g}}_k$. These VRS~gradients~$\textvec{\ddot{g}}_k$ are used in the coordinate-wise aggregation step in Eq. \ref{eq:fedrise-alignaggr} instead of the original client gradients $\textvec{g}_k$. Hence, we refer to the complete aggregation procedure as \emph{robust} coordinate-wise aggregation (RoCA).

It must be emphasized that, in our experiments, the CRISE step uses the original gradients $\textvec{g}_k$ to compute the concordance ratio, whereas the sign election uses the
sparsified gradients ($\textvec{\starhat{g}}_k \leftarrow \textvec{\ddot{g}}_k$).


\paragraph{Server Momentum}: Finally, we add an \textit{optional} momentum term to include updates from the previous round. The momentum term can smooth out the noise introduced by the Byzantines that has not been filtered out. More importantly, it ensures the updates do not become zero when the rating of all clients diverges due to statistical heterogeneity, causing no sign election. Momentum will ensure that updates will continue to move in a specific direction based on previous updates, thus moving out of the saddle point.

\begin{equation}
\begin{aligned}
    \widebreve{\textvec{g}}^{(t)} = \beta_\texttt{ra}\cdot \widebreve{\textvec{g}}^{(t-1)} + (1-\beta_\texttt{ra})\cdot\widetilde{\textvec{g}}^{(t)} \\
\end{aligned}
\label{eq:fedrise-momagg}
\end{equation}

\noindent Either $\widebreve{\textvec{g}}^{(t)}$ or $\widetilde{\textvec{g}}^{(t)}$ can be used to update the global model, depending on whether server momentum is applied or not.



\begin{figure*}[h]
\begin{center}
\includegraphics[width=0.31\linewidth]{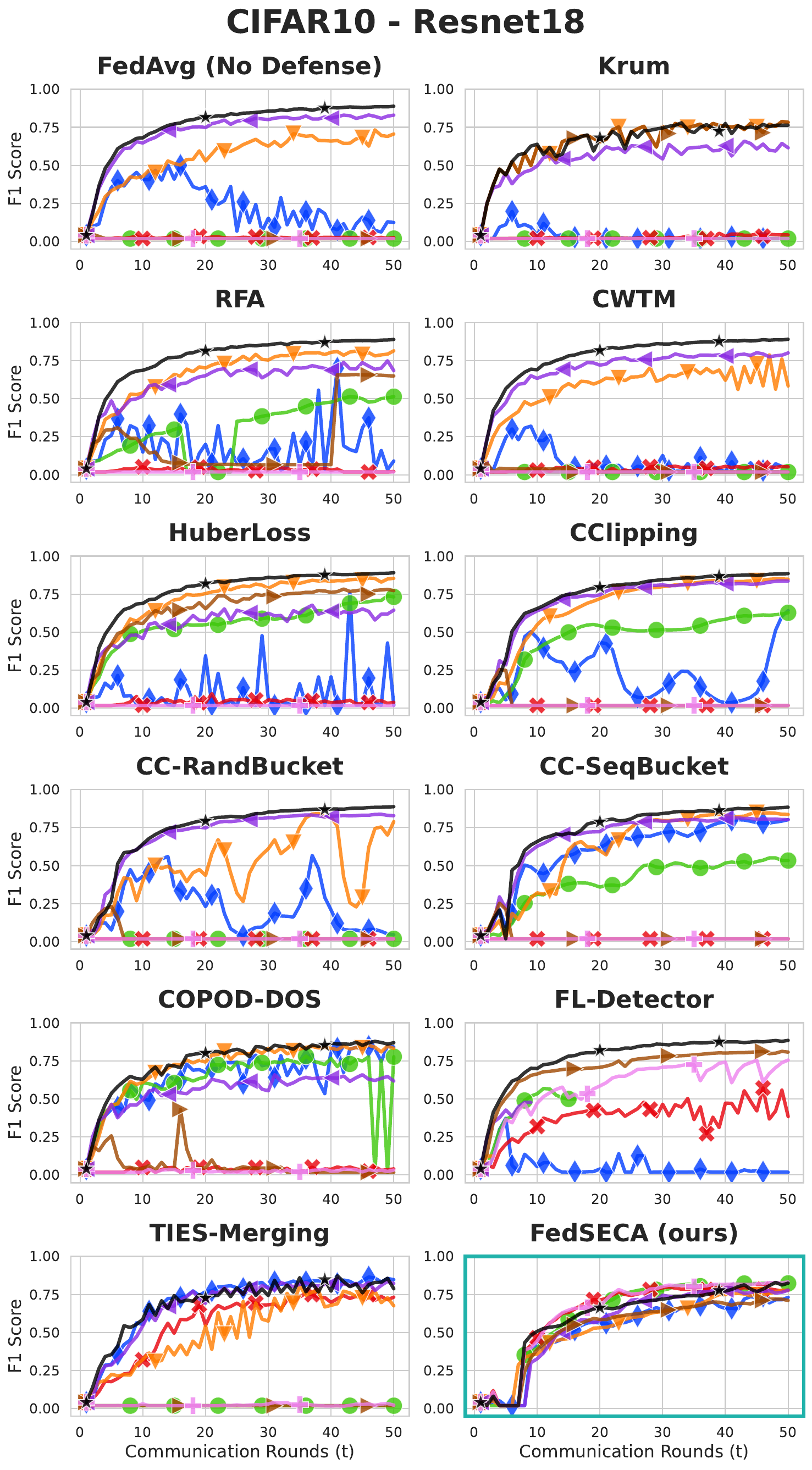}
\hspace{1ex}
\includegraphics[width=0.31\linewidth]{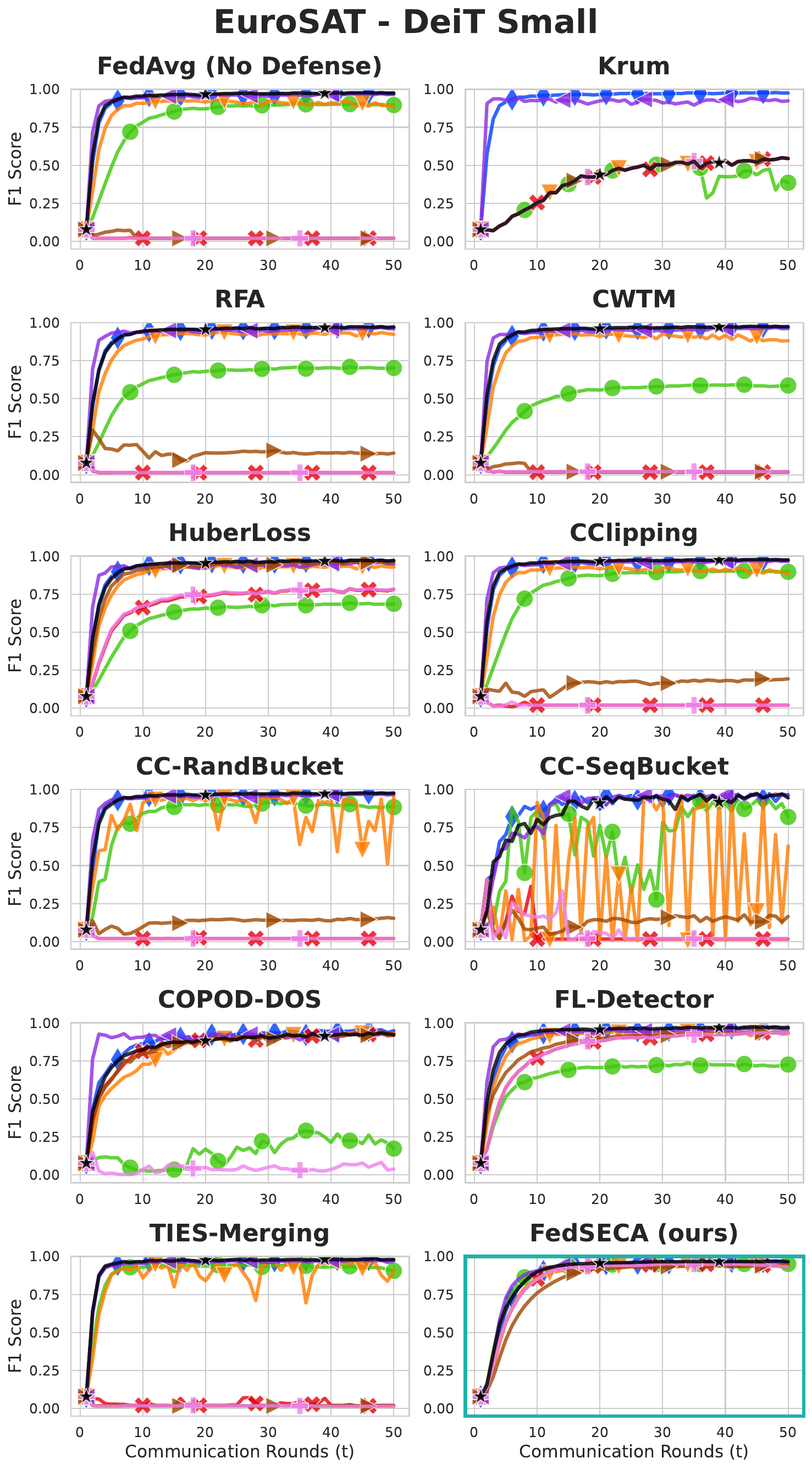}
\hspace{1ex}
\includegraphics[width=0.31\linewidth]{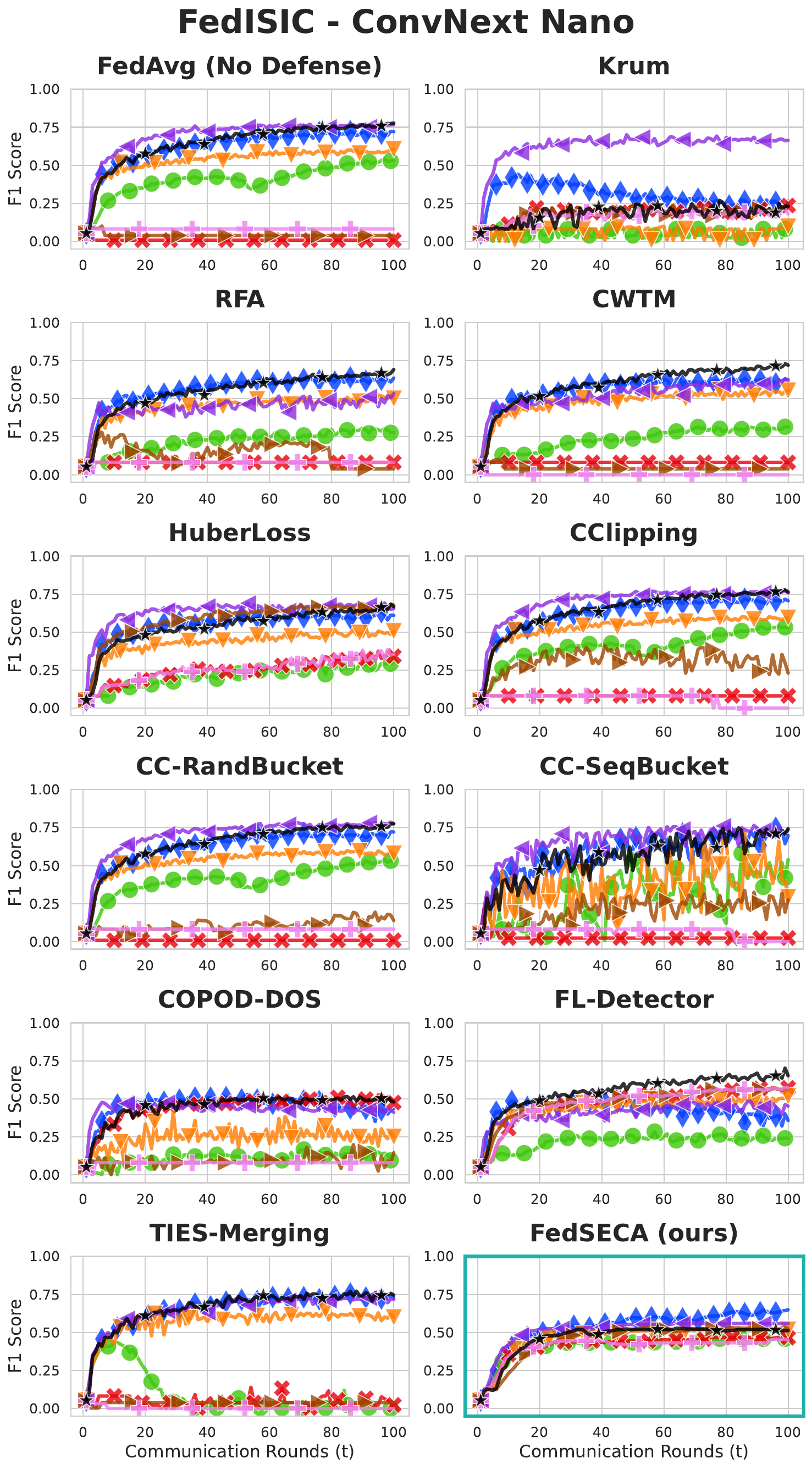}
\includegraphics[width=0.55\linewidth]{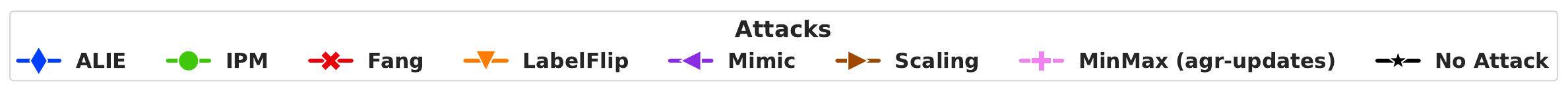}

\end{center}
   \caption{\textbf{\textit{Performance comparison of $\textaggfunc{RAggr}$ methods}}:
    Each subplot corresponds to a $\textaggfunc{RAggr}$ method, and colored line-plots within subplots correspond to specific attacks. 
    The line plot indicates test F1-scores computed on a pooled test set at the end of each round $t$. Analyzing the performance trajectory is essential for assessing the stability of a $\textaggfunc{RAggr}$ function, as focusing only on final test accuracy will be misleading. While FedSECA robustly converges to a good performance for all attacks, other methods fail for some attacks.}
\label{fig:results-all}
\end{figure*}


\section{Experiments} \label{sec:experiments}

\subsection{Performance Comparison of \textaggfunc{RAggr} Methods}
\noindent \textbf{Baselines}: We compare the proposed FedSECA approach against 10 baseline \textaggfunc{RAggr} methods. These include middle-seeking methods such as \textit{Krum} \cite{defns-blanchard2017machine_krum}, \textit{RFA} \cite{defns-pillutla2022rfa_geomed}, and coordinate-wise trimmed mean (\textit{CWTM}) \cite{defns-yin2018byzantine_cwtm}, Huber loss minimization in aggregator (\textit{HuberLoss}) \cite{defns-zhao2024huberloss}, clipping \& bucketing strategies such as Centered Clipping (\textit{CClipping}) \cite{defns-karimireddy2021learning_clipping}, clipping with random bucketing (\textit{CC-RandBucket}) \cite{defns-karimireddy2020byzantine_randBuck}, and clipping with sequential bucketing (\textit{CC-SeqBucket}) \cite{defns-ozfatura2023byzantines_seqbuck}, statistical detection methods such as \textit{COPOD-DOS} \cite{defns-alkhunaizi2022suppressing_copod} and \textit{FL-Detector} \cite{defns-zhang2022fldetector}. Finally, to assess the effectiveness of vanilla sign-based merging, we also compare against modified \textit{TIES-Merging} \cite{merge-yadav2023resolving_ties}, where sign election is changed from gradient magnitude summation to sign summation across clients for a given coordinate. The results for vanilla \textit{FedAvg} (no defense) are also reported for comparison. 
\textbf{Attacks}: We evaluate the effectiveness of the \textaggfunc{RAggr} methods based on 7 attacks,  \textit{ALIE} \cite{attk-baruch2019little}, \textit{IPM} \cite{attk-xie2020fall}, \textit{Fang} \cite{attk-fang2020local}, \textit{LabelFlip} \cite{attk-biggio2012poisoning_labelflip}, \textit{Mimic} \cite{defns-karimireddy2020byzantine_randBuck}, \textit{Scaling} and \textit{MinMax (agr-updates)}\cite{attk-hejwalkar2021minmax_minsum_dnc}. We also consider the \textit{No Attack} scenario to assess the impact of the \textaggfunc{RAggr} methods on the model performance in the absence of any Byzantine clients.
\textbf{Datasets}: Benchmarking of the above methods is conducted on 3 image classification datasets from different domains, and a different model architecture is used for each domain. The goal is to evaluate effectiveness of $\textaggfunc{RAggr}$ methods under three scenarios: \textit{\textbf{(a)}~Training from scratch with heterogeneity:} ResNet18 \cite{model-he2016resnet} model is trained on CIFAR10 \cite{data-krizhevsky2009learning_cifar} dataset (10 classes), which is split into 5 heterogeneous clients following a Dirichlet distribution with parameter $1$. 
\textit{\textbf{(b)} Finetuning a pre-trained model using heterogeneous data with good inter-class separability:} DeiT-S model \cite{model-touvron2021training_deit} pre-trained on ImageNet is finetuned on EuroSAT \cite{data-helber2019eurosat} dataset (10 classes), which is split into 7 clients based on geographical regions.
\textit{\textbf{(c)} Finetuning a pre-trained model using heterogeneous data with poor inter-class separability:} ConvNeXt-nano model \cite{model-woo2023convnext} pre-trained on ImageNet is finetuned on FedISIC \cite{data-ogier2022flamby} dataset (8 melanoma classes), which is split into 6 clients based on hospitals. In all three settings, the number of Byzantines $\textcount{B} = 2$ and the same training hyperparameters are used. Details of all the hyperparameters involved in training, attacks, and defense methods are discussed in the supplementary material. The hyperparameters of FedSECA, namely the sparsification factor $\gamma=0.9$ and server momentum $\beta_\texttt{ra}=0.5$ are also kept the same for all three datasets.


Our main results in \cref{fig:results-all} lead to the following observations. (i) Most $\textaggfunc{RAggr}$ methods fail against Fang, Scaling and MinMax attacks, right from the start of collaboration. (ii) Since ALIE attack slowly diverges the training, prolonged exposure to this attack causes instability in many defenses, especially when training from scratch. (iii) IPM derails most robust aggregators when trained from scratch, but does not have as much impact when finetuning pre-trained models for simpler tasks (EuroSAT). This is because very small updates happen in each training step, making it hard to diverge. 
(iv)~The impact of Mimic and LabelFlip attacks is minimal in most cases. 
Finally, \textit{FedSECA successfully resists all of the attacks} and converges to an optimal solution on both CIFAR10 and EuroSAT datasets. In FedISIC, FedSECA converges to a suboptimal solution, resulting in lower accuracy compared to FedAvg. This is due to the challenging nature of the task/dataset, which requires fine-grained visual classification of skin melanoma. We posit that FedSECA is marginally more cautious than some baselines and eliminates finer gradients, which may be critical for better separability among classes in challenging tasks. However, even in this setting, FedSECA performs better on average across all attacks compared to the baselines. 



\subsection{Additional Results}

\begin{table}[t]
\centering
\resizebox{\linewidth}{!}{
\setlength{\tabcolsep}{4pt}
\begin{tabular}{l|l| ccccccc |c}
\toprule
    Components & &
    \rotatebox{90}{ALIE} & \rotatebox{90}{IPM} & \rotatebox{90}{Fang} & 
    \rotatebox{90}{LabelFlip} & \rotatebox{90}{Mimic} & \rotatebox{90}{Scaling} & 
    \rotatebox{90}{No Attack} & Mean \\
\midrule
FedSECA  (Full)                &
& 0.72 &  0.82 &  0.78 &  0.76 &  0.77 &  0.72 &  0.81                         &\underline{0.77}\\

\without Concordance Ratio     &$\rho_{k}=1$
&  0.72 &  \cellcolor{Tred}0.02 &  0.81 &  \cellcolor{Tyellow}0.67 &  0.80 &  0.72 &  0.78        &0.65\\

\without Coord-wise Aggr.   &$\delta_{k}^{j} = \rho_{k}$
&  \cellcolor{Torange}0.23 &  0.77 &  0.80 &  \cellcolor{Tyellow}0.57 &  0.75 & \cellcolor{Tyellow} 0.63 &  \cellcolor{Tyellow}0.68     &0.63\\

\without CRISE                 & $\delta_{k}^{j} = 1$
&  0.75 & \cellcolor{Tred}0.03 &  0.81 &  \cellcolor{Tyellow}0.67 &  0.81 &  0.73 &  0.77         &0.65\\

\without VRS Gradients         & $\textvec{\starhat{g}}_{k} = \textvec{g}_{k}$
&  \cellcolor{Tred}0.03 & 0.83 &  \cellcolor{Torange}0.34 &  0.85 &  \cellcolor{Tyellow}0.67 &  \cellcolor{Tred}0.02 &  0.88        &0.52\\

\without Sparsification        & $\gamma = 0$
&  \cellcolor{Tred}0.02 &  0.83 &  0.80 &  0.81 &  \cellcolor{Tyellow}0.63 &  0.79 &  0.87        &0.68\\

\without Clip+Clamp            & $\tau,\mu^j=\infty$
&  0.85 &  0.81 &  0.80 &  0.81 &  0.83 &  \cellcolor{Tred}0.03 &  0.85        &0.71\\

\without Clipping              & $\tau = \infty$
&  0.71 &  0.82 &  0.81 &  0.79 &  0.79 &  \cellcolor{Tred}0.02 &  0.79        &0.68\\

\without Clamping              & $\mu^j = \infty$
&  0.85 &  0.81 &  0.80 &  0.83 &  0.82 &  \cellcolor{Torange}0.17 &  0.85     &0.73\\

\without Server Momentum       & $\beta_\texttt{ra} = 0$
&  0.70 &  0.85 &  0.82 &  0.79 &  0.79 &  0.74 &  0.81                        &\textbf{0.79}\\

\bottomrule
\end{tabular}
}
  \caption{\textbf{\textit{Ablation study}} showing the relative impact of each component in the FedSECA pipeline. Here, \colorbox{Tred}{\footnotesize red} indicates that the training collapsed due to the attack, \colorbox{Torange}{\footnotesize orange} indicates that the training was impacted severely, and \colorbox{Tyellow}{\footnotesize yellow} indicates a noticeable drop in accuracy. Values in this table correspond to the mean F1-score of the last 5 communication rounds computed on the test set.}
  \label{tab:fedrise_components}
\end{table}

\begin{figure}[h]
\begin{center}
\includegraphics[width=\linewidth]{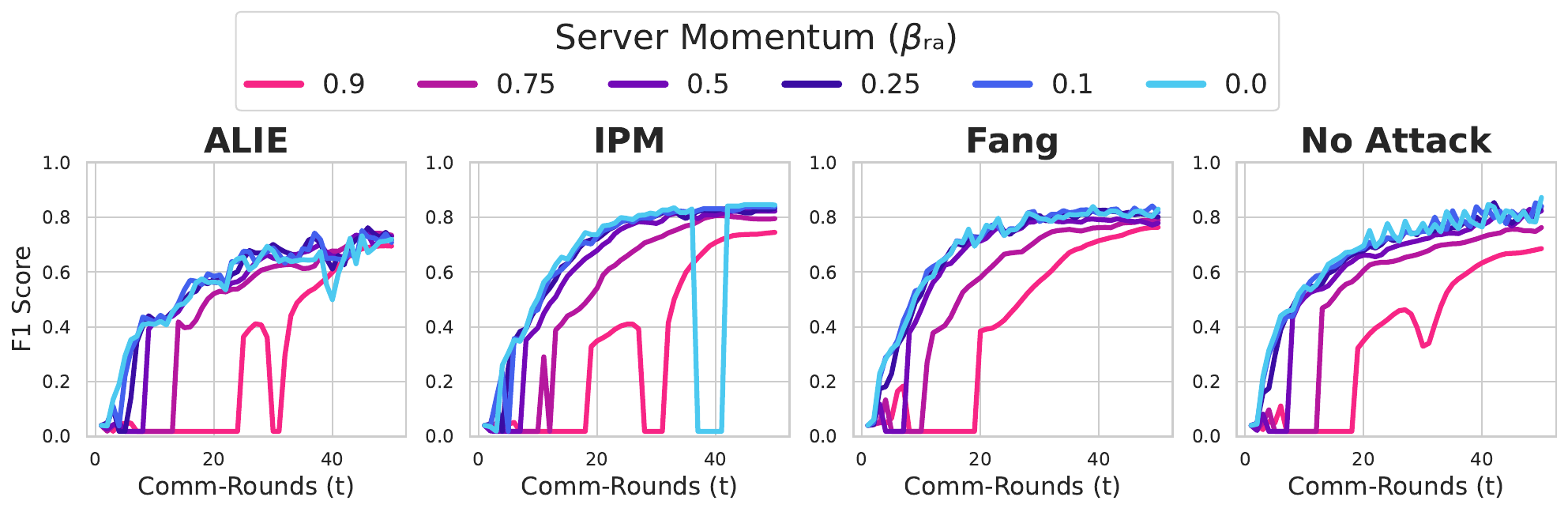}
\includegraphics[width=\linewidth]{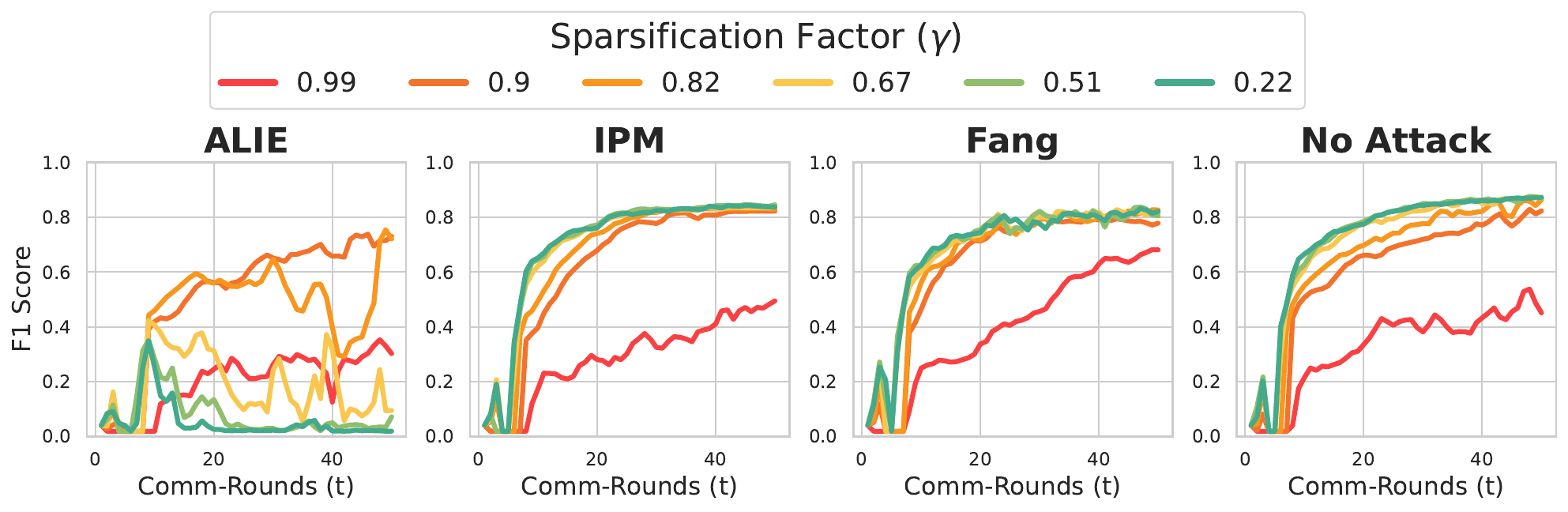}
\end{center}
   \caption{\textbf{\textit{Hyperparameter Sensitivity of FedSECA method}}. While the top row shows sensitivity to server momentum $\beta_\texttt{ra}$, the bottom row shows sensitivity to sparsification factor $\gamma$.}
\label{fig:ablate-hypers}
\end{figure}

\begin{figure}[t]
\begin{center}
\includegraphics[width=\linewidth]{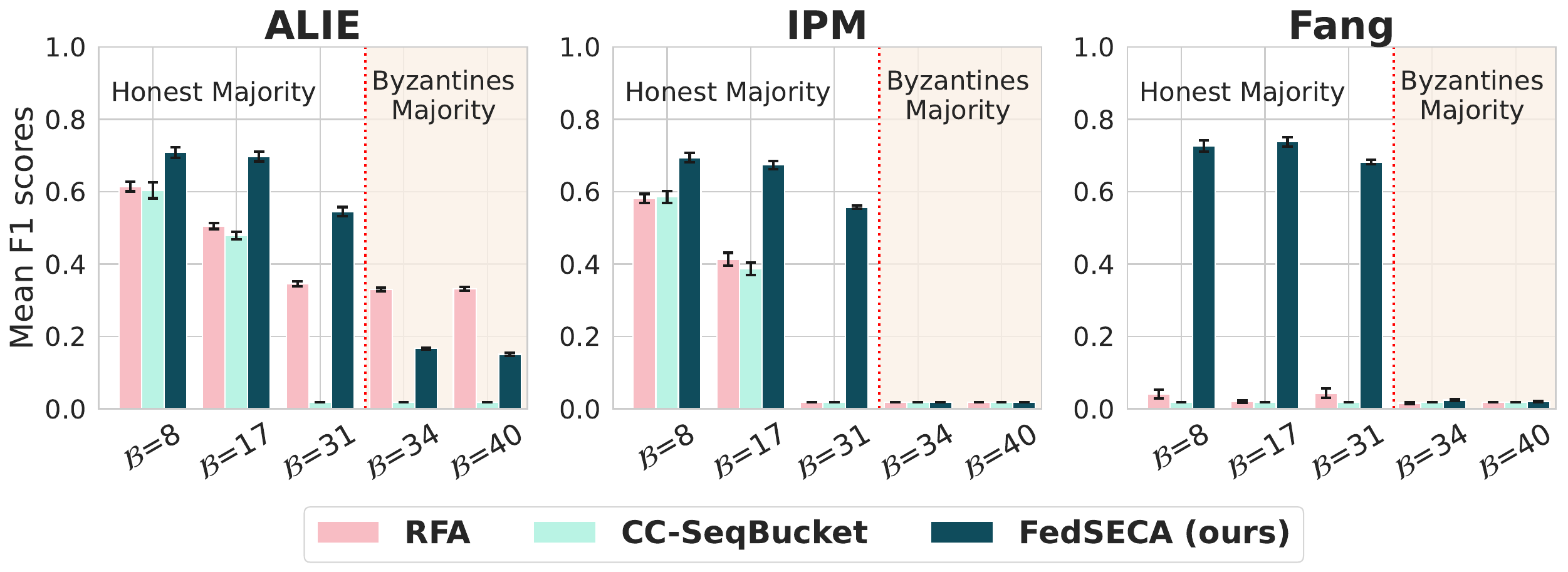}
\end{center}
   \caption{\textbf{\textit{Robustness of FedSECA to number of Byzantine clients}}~($\textcount{B}$). 
   X-axis represents different $\textcount{B}$ values with fixed $\textcount{K} = 64$. 
   }
\label{fig:ansys-byzantcount}
\end{figure}

\noindent \textbf{Ablation Study}: To understand the impact of various components within FedSECA, we conduct an ablation study based on the same CIFAR10 split as explained earlier. From \cref{tab:fedrise_components}, we observe that the entire FedSECA pipeline gives the best average performance across all attacks.  While the momentum term may seem unhelpful, it prevents updates from getting stuck when all $\rho_k$ become zero under extreme heterogeneity. Hence, we consider momentum as optional.

\noindent \textbf{Sensitivity to Hyperparameters}: Sensitivity of FedSECA to hyperparameters $\gamma$ and $\beta_{\texttt{ra}}$ is evaluated in \cref{fig:ablate-hypers} based on the same CIFAR10 split. Convergence is unaffected by a range of $\beta_\texttt{ra}$, except for large values ($0.9$). In general, higher momentum leads to poor training stability because sparse gradients quickly become stale. High sparsity ($\gamma=0.99$) harms convergence because a substantial number of updates are ignored and the training struggles to recover. Performance under the ALIE attack is notably sensitive to $\gamma$ because perturbations in this attack have low magnitude, and a lower value of $\gamma$ includes many corrupted updates. 


   
\noindent \textbf{Robustness to number of Byzantine clients}: For this study, we use CIFAR10 dataset with $\textcount{K} = 64$ clients and vary the number of Byzantines $\textcount{B}$. From Figure \cref{fig:ansys-byzantcount}, we observe that FedSECA is stable even with an increasing proportion of Byzantines and especially performs very well under the Fang attack. Our method is resilient even at 48\% Byzantines and collapses only when Byzantines increase beyond $50\%$, which is expected for any \textaggfunc{RAggr}. 



\begin{table}[h]
\centering
\resizebox{\linewidth}{!}{
\begin{tabular}{ccccccc}
\toprule
\midrule
 No. Clients & \textbf{5 clients} & \textbf{10 clients} & \textbf{25 clients} & \textbf{50 clients} & \textbf{100 clients} & \textbf{200 clients} \\
\midrule
Images/Client & 10K & 5K & 2K & 1K & 500 & 250 \\
\midrule
\midrule
FedAvg  & {89.1}$_{\pm(0.3)}$ & {85.8}$_{\pm(0.7)}$ & {75.9}$_{\pm(1.0)}$ & {67.4}$_{\pm(1.3)}$ & {57.0}$_{\pm(1.1)}$ & {48.1}$_{\pm(1.3)}$  \\
\midrule
FedSECA  & {87.2}$_{\pm(0.5)}$ & {83.1}$_{\pm(1.0)}$ & {76.8}$_{\pm(1.6)}$ & {69.7}$_{\pm(1.4)}$ & {63.0}$_{\pm(1.4)}$ & {55.3}$_{\pm(2.2)}$ \\
\midrule
\bottomrule
\end{tabular}
}
\caption{\textbf{\textit{Scalability of FedSECA to the number of clients in FL.}}}
\label{tab:ansys_client_count}
\end{table}

\noindent \textbf{Scalability of FedSECA with number of clients}: We analyze the scalability of FedSECA using CIFAR10 IID split with varying client count $\textcount{K}$ in \cref{tab:ansys_client_count}. Note that FedSECA is designed for cross-silo FL applications, where the number of clients is expected to be less than 100 \cite{fed-kairouz2021advances}. We train for 50 epochs, keeping the training setting the same across all client counts, and we report the mean score of the last 10 rounds. This scalability evaluation is conducted under a \textbf{\textit{no-attack setting}} and compares FedSECA against FedAvg since it provides the upper bound for accuracy when all clients are honest. It can be seen that FedSECA is closer to FedAvg and even exceeds it when the client count increases because FedSECA effectively filters out noisy updates.


\section{Conclusion}

We introduced a robust aggregation method called FedSECA for Byzantine-tolerant FL. Since FedSECA averages only those gradients that align with the sign consensus along each coordinate, it converges for all the attack settings considered in this work. The main limitation of FedSECA is the quadratic computational complexity when the number of clients ($\textcount{K}$) increases, which limits its applicability to cross-device FL scenarios (with thousands of clients). Client selection strategies could be used to address this issue. In future, FedSECA could be integrated as part of a dynamic framework, wherein individual components could be enabled based on training stability using game theory \cite{defns-xie2023mixed_nash}. This could ensure optimal convergence in the absence of Byzantines and robust aggregation under severe attacks.

\paragraph{Acknowledgement.}
This material is based in part on work supported by the Office of Naval Research N00014-24-1-2168.

{
    \small
    \bibliographystyle{ieeenat_fullname}
    \bibliography{main_reference}

@String(ICASSP=	{ICASSP})

@String(AAAI = {AAAI})

@article{data-ogier2022flamby,
  title={Flamby: Datasets and benchmarks for cross-silo federated learning in realistic healthcare settings},
  author={Ogier du Terrail, Jean and Ayed, Samy-Safwan and Cyffers, Edwige and Grimberg, Felix and He, Chaoyang and Loeb, Regis and Mangold, Paul and Marchand, Tanguy and Marfoq, Othmane and Mushtaq, Erum and others},
  journal={Advances in Neural Information Processing Systems},
  volume={35},
  pages={5315--5334},
  year={2022}
}

@article{data-krizhevsky2009learning_cifar,
  title={Learning multiple layers of features from tiny images},
  author={Krizhevsky, Alex and Hinton, Geoffrey and others},
  year={2009},
  publisher={Toronto, ON, Canada}
}

@article{data-helber2019eurosat,
  title={Eurosat: A novel dataset and deep learning benchmark for land use and land cover classification},
  author={Helber, Patrick and Bischke, Benjamin and Dengel, Andreas and Borth, Damian},
  journal={IEEE Journal of Selected Topics in Applied Earth Observations and Remote Sensing},
  volume={12},
  number={7},
  pages={2217--2226},
  year={2019},
  publisher={IEEE}
}

@inproceedings{model-he2016resnet,
  title={Deep residual learning for image recognition},
  author={He, Kaiming and Zhang, Xiangyu and Ren, Shaoqing and Sun, Jian},
  booktitle={Proceedings of the IEEE conference on computer vision and pattern recognition},
  pages={770--778},
  year={2016}
}

@inproceedings{model-touvron2021training_deit,
  title={Training data-efficient image transformers \& distillation through attention},
  author={Touvron, Hugo and Cord, Matthieu and Douze, Matthijs and Massa, Francisco and Sablayrolles, Alexandre and J{\'e}gou, Herv{\'e}},
  booktitle={International conference on machine learning},
  pages={10347--10357},
  year={2021},
  organization={PMLR}
}

@inproceedings{model-woo2023convnext,
  title={Convnext v2: Co-designing and scaling convnets with masked autoencoders},
  author={Woo, Sanghyun and Debnath, Shoubhik and Hu, Ronghang and Chen, Xinlei and Liu, Zhuang and Kweon, In So and Xie, Saining},
  booktitle={Proceedings of the IEEE/CVF Conference on Computer Vision and Pattern Recognition},
  pages={16133--16142},
  year={2023}
}

@article{merge-lin2017deepcompression,
  title={Deep gradient compression: Reducing the communication bandwidth for distributed training},
  author={Lin, Yujun and Han, Song and Mao, Huizi and Wang, Yu and Dally, William J},
  journal={arXiv preprint arXiv:1712.01887},
  year={2017}
}

@article{merge-yadav2023resolving_ties,
  title={Resolving interference when merging models},
  author={Yadav, Prateek and Tam, Derek and Choshen, Leshem and Raffel, Colin and Bansal, Mohit},
  journal={arXiv preprint arXiv:2306.01708},
  year={2023}
}

@inproceedings{fed-mcmahan2017communication,
  title={Communication-efficient learning of deep networks from decentralized data},
  author={McMahan, Brendan and Moore, Eider and Ramage, Daniel and Hampson, Seth and y Arcas, Blaise Aguera},
  booktitle={Artificial intelligence and statistics},
  pages={1273--1282},
  year={2017},
  organization={PMLR}
}

@article{fed-kairouz2021advances,
  title={Advances and open problems in federated learning},
  author={Kairouz, Peter and McMahan, H Brendan and others},
  journal={Foundations and trends{\textregistered} in machine learning},
  volume={14},
  number={1--2},
  pages={1--210},
  year={2021},
  publisher={Now Publishers, Inc.}
}

@article{attk_jere2020taxonomy,
  title={A taxonomy of attacks on federated learning},
  author={Jere, Malhar S and Farnan, Tyler and Koushanfar, Farinaz},
  journal={IEEE Security \& Privacy},
  volume={19},
  number={2},
  pages={20--28},
  year={2020},
  publisher={IEEE}
}

@inproceedings{attk-fang2020local,
  title={Local model poisoning attacks to $\{$Byzantine-Robust$\}$ federated learning},
  author={Fang, Minghong and Cao, Xiaoyu and Jia, Jinyuan and Gong, Neil},
  booktitle={29th USENIX security symposium (USENIX Security 20)},
  pages={1605--1622},
  year={2020}
}

@article{attk-baruch2019little,
  title={A Little Is Enough: Circumventing Defenses For Distributed Learning},
  author={Baruch, Gilad and Baruch, Moran and Goldberg, Yoav},
  journal={Advances in Neural Information Processing Systems},
  volume={32},
  year={2019}
}

@inproceedings{attk-xie2020fall,
  title={Fall of empires: Breaking byzantine-tolerant sgd by inner product manipulation},
  author={Xie, Cong and Koyejo, Oluwasanmi and Gupta, Indranil},
  booktitle={Uncertainty in Artificial Intelligence},
  pages={261--270},
  year={2020},
  organization={PMLR}
}

@inproceedings{attk-biggio2012poisoning_labelflip,
  title={Poisoning attacks against support vector machines},
  author={Biggio, Battista and Nelson, Blaine and Laskov, Pavel},
  booktitle={Proceedings of the 29th International Coference on International Conference on Machine Learning},
  pages={1467--1474},
  year={2012}
}

@inproceedings{attk-hejwalkar2021minmax_minsum_dnc,
  title={Manipulating the byzantine: Optimizing model poisoning attacks and defenses for federated learning},
  author={Shejwalkar, Virat and Houmansadr, Amir},
  booktitle={NDSS},
  year={2021}
}

@article{attk-xie2024poisonedfl,
  title={PoisonedFL: Model Poisoning Attacks to Federated Learning via Multi-Round Consistency},
  author={Xie, Yueqi and Fang, Minghong and Gong, Neil Zhenqiang},
  journal={arXiv preprint arXiv:2404.15611},
  year={2024}
}

@article{defns-blanchard2017machine_krum,
  title={Machine learning with adversaries: Byzantine tolerant gradient descent},
  author={Blanchard, Peva and El Mhamdi, El Mahdi and Guerraoui, Rachid and Stainer, Julien},
  journal={Advances in neural information processing systems},
  volume={30},
  year={2017}
}

@inproceedings{defns-yin2018byzantine_cwtm,
  title={Byzantine-robust distributed learning: Towards optimal statistical rates},
  author={Yin, Dong and Chen, Yudong and Kannan, Ramchandran and Bartlett, Peter},
  booktitle={International Conference on Machine Learning},
  pages={5650--5659},
  year={2018},
  organization={Pmlr}
}

@inproceedings{defns-alkhunaizi2022suppressing_copod,
  title={Suppressing poisoning attacks on federated learning for medical imaging},
  author={Alkhunaizi, Naif and Kamzolov, Dmitry and Tak{\'a}{\v{c}}, Martin and Nandakumar, Karthik},
  booktitle={International Conference on Medical Image Computing and Computer-Assisted Intervention},
  pages={673--683},
  year={2022},
  organization={Springer}
}

@article{defns-pillutla2022rfa_geomed,
  title={Robust aggregation for federated learning},
  author={Pillutla, Krishna and Kakade, Sham M and Harchaoui, Zaid},
  journal={IEEE Transactions on Signal Processing},
  volume={70},
  pages={1142--1154},
  year={2022},
  publisher={IEEE}
}

@inproceedings{defns-karimireddy2021learning_clipping,
  title={Learning from history for byzantine robust optimization},
  author={Karimireddy, Sai Praneeth and He, Lie and Jaggi, Martin},
  booktitle={International Conference on Machine Learning},
  pages={5311--5319},
  year={2021},
  organization={PMLR}
}

@article{defns-karimireddy2020byzantine_randBuck,
  title={Byzantine-robust learning on heterogeneous datasets via bucketing},
  author={Karimireddy, Sai Praneeth and He, Lie and Jaggi, Martin},
  journal={arXiv preprint arXiv:2006.09365},
  year={2020}
}

@article{defns-ozfatura2023byzantines_seqbuck,
  title={Byzantines can also learn from history: Fall of centered clipping in federated learning},
  author={{\"O}zfatura, Kerem and {\"O}zfatura, Emre and K{\"u}p{\c{c}}{\"u}, Alptekin and G{\"u}nd{\"u}z, Deniz},
  journal={IEEE Transactions on Information Forensics and Security},
  year={2023},
  publisher={IEEE}
}

@InProceedings{defns-xu2020reputation_RFFL,
    title={A Reputation Mechanism Is All You Need: Collaborative Fairness and Adversarial Robustness in Federated Learning},
    author={Xinyi Xu and Lingjuan Lyu},
    year={2021},
    booktitle={International Workshop on Federated Learning for User Privacy and Data Confidentiality in Conjunction with ICML 2021 (FL-ICML'21)},
}

@article{defns-bernstein2018signsgd,
  title={signSGD with majority vote is communication efficient and fault tolerant},
  author={Bernstein, Jeremy and Zhao, Jiawei and Azizzadenesheli, Kamyar and Anandkumar, Anima},
  journal={arXiv preprint arXiv:1810.05291},
  year={2018}
}

@article{defns-sohn2020election_coding_signsgd,
  title={Election coding for distributed learning: Protecting signsgd against byzantine attacks},
  author={Sohn, Jy-yong and Han, Dong-Jun and Choi, Beongjun and Moon, Jaekyun},
  journal={Advances in Neural Information Processing Systems},
  volume={33},
  pages={14615--14625},
  year={2020}
}

@article{defns-mhamdi2018hidden_bulyan,
  title={The hidden vulnerability of distributed learning in byzantium},
  author={Mhamdi, El Mahdi El and Guerraoui, Rachid and Rouault, S{\'e}bastien},
  journal={arXiv preprint arXiv:1802.07927},
  year={2018}
}

@inproceedings{defns-allouah2023fixingbymixing,
  title={Fixing by mixing: A recipe for optimal byzantine ml under heterogeneity},
  author={Allouah, Youssef and Farhadkhani, Sadegh and Guerraoui, Rachid and Gupta, Nirupam and Pinot, Rafa{\"e}l and Stephan, John},
  booktitle={International Conference on Artificial Intelligence and Statistics},
  pages={1232--1300},
  year={2023},
  organization={PMLR}
}

@inproceedings{defns-xu2022byzantine_signguard,
  title={Byzantine-robust federated learning through collaborative malicious gradient filtering},
  author={Xu, Jian and Huang, Shao-Lun and Song, Linqi and Lan, Tian},
  booktitle={2022 IEEE 42nd International Conference on Distributed Computing Systems (ICDCS)},
  pages={1223--1235},
  year={2022},
  organization={IEEE}
}

@article{defns-li2023experimental_clippedclustering,
  title={An experimental study of byzantine-robust aggregation schemes in federated learning},
  author={Li, Shenghui and Ngai, Edith C-H and Voigt, Thiemo},
  journal={IEEE Transactions on Big Data},
  year={2023},
  publisher={IEEE}
}

@inproceedings{defns-sattler2020byzantine_cosineclustering,
  title={On the byzantine robustness of clustered federated learning},
  author={Sattler, Felix and M{\"u}ller, Klaus-Robert and Wiegand, Thomas and Samek, Wojciech},
  booktitle={ICASSP 2020-2020 IEEE International Conference on Acoustics, Speech and Signal Processing (ICASSP)},
  pages={8861--8865},
  year={2020},
  organization={IEEE}
}

@article{defns-rajput2019detox_robust,
  title={DETOX: A redundancy-based framework for faster and more robust gradient aggregation},
  author={Rajput, Shashank and Wang, Hongyi and Charles, Zachary and Papailiopoulos, Dimitris},
  journal={Advances in Neural Information Processing Systems},
  volume={32},
  year={2019}
}

@article{defns-xie2023mixed_nash,
  title={Mixed Nash for Robust Federated Learning},
  author={Xie, Wanyun and Pethick, Thomas and Ramezani-Kebrya, Ali and Cevher, Volkan},
  journal={Transactions on Machine Learning Research},
  year={2023}
}

@article{defns-zhang2022flip_datapoisoning,
  title={Flip: A provable defense framework for backdoor mitigation in federated learning},
  author={Zhang, Kaiyuan and Tao, Guanhong and Xu, Qiuling and Cheng, Siyuan and An, Shengwei and Liu, Yingqi and Feng, Shiwei and Shen, Guangyu and Chen, Pin-Yu and Ma, Shiqing and others},
  journal={arXiv preprint arXiv:2210.12873},
  year={2022}
}

@inproceedings{defns-zhang2022fldetector,
  title={Fldetector: Defending federated learning against model poisoning attacks via detecting malicious clients},
  author={Zhang, Zaixi and Cao, Xiaoyu and Jia, Jinyuan and Gong, Neil Zhenqiang},
  booktitle={Proceedings of the 28th ACM SIGKDD conference on knowledge discovery and data mining},
  pages={2545--2555},
  year={2022}
}

@inproceedings{defns-zhao2024huberloss,
  title={A huber loss minimization approach to byzantine robust federated learning},
  author={Zhao, Puning and Yu, Fei and Wan, Zhiguo},
  booktitle={Proceedings of the AAAI Conference on Artificial Intelligence},
  volume={38},
  number={19},
  pages={21806--21814},
  year={2024}
}

@article{defns-xu2024LASAByzantine,
  title={Achieving Byzantine-Resilient Federated Learning via Layer-Adaptive Sparsified Model Aggregation},
  author={Xu, Jiahao and Zhang, Zikai and Hu, Rui},
  journal={arXiv preprint arXiv:2409.01435},
  year={2024}
}

@article{lit-nguyen2024backdoor,
  title={Backdoor attacks and defenses in federated learning: Survey, challenges and future research directions},
  author={Nguyen, Thuy Dung and Nguyen, Tuan and Le Nguyen, Phi and Pham, Hieu H and Doan, Khoa D and Wong, Kok-Seng},
  journal={Engineering Applications of Artificial Intelligence},
  volume={127},
  pages={107166},
  year={2024},
  publisher={Elsevier}
}

@inproceedings{lit-tolpegin2020datapoison,
  title={Data poisoning attacks against federated learning systems},
  author={Tolpegin, Vale and Truex, Stacey and Gursoy, Mehmet Emre and Liu, Ling},
  booktitle={Computer Security--ESORICS 2020: 25th European Symposium on Research in Computer Security, ESORICS 2020, Guildford, UK, September 14--18, 2020, Proceedings, Part I 25},
  pages={480--501},
  year={2020},
  organization={Springer}
}

@article{lit-nair2023robust_adversarial,
  title={A robust analysis of adversarial attacks on federated learning environments},
  author={Nair, Akarsh K and Raj, Ebin Deni and Sahoo, Jayakrushna},
  journal={Computer Standards \& Interfaces},
  volume={86},
  pages={103723},
  year={2023},
  publisher={Elsevier}
}
}



\cleardoublepage 
\renewcommand\thesection{\Alph{section}}
\renewcommand\thefigure{\thesection.\arabic{figure}}
\renewcommand\thetable{\thesection.\arabic{table}}
\setcounter{figure}{0}
\setcounter{table}{0}
\setcounter{section}{0}

\begin{table}[t]
    \centering
    {\LARGE \textbf{Supplementary Material}}
\end{table}

\maketitle

\section{Data Distributions} \label{sup-sec:data_dist}
\begin{figure}[h]
\begin{center}
\includegraphics[width=0.75\linewidth]{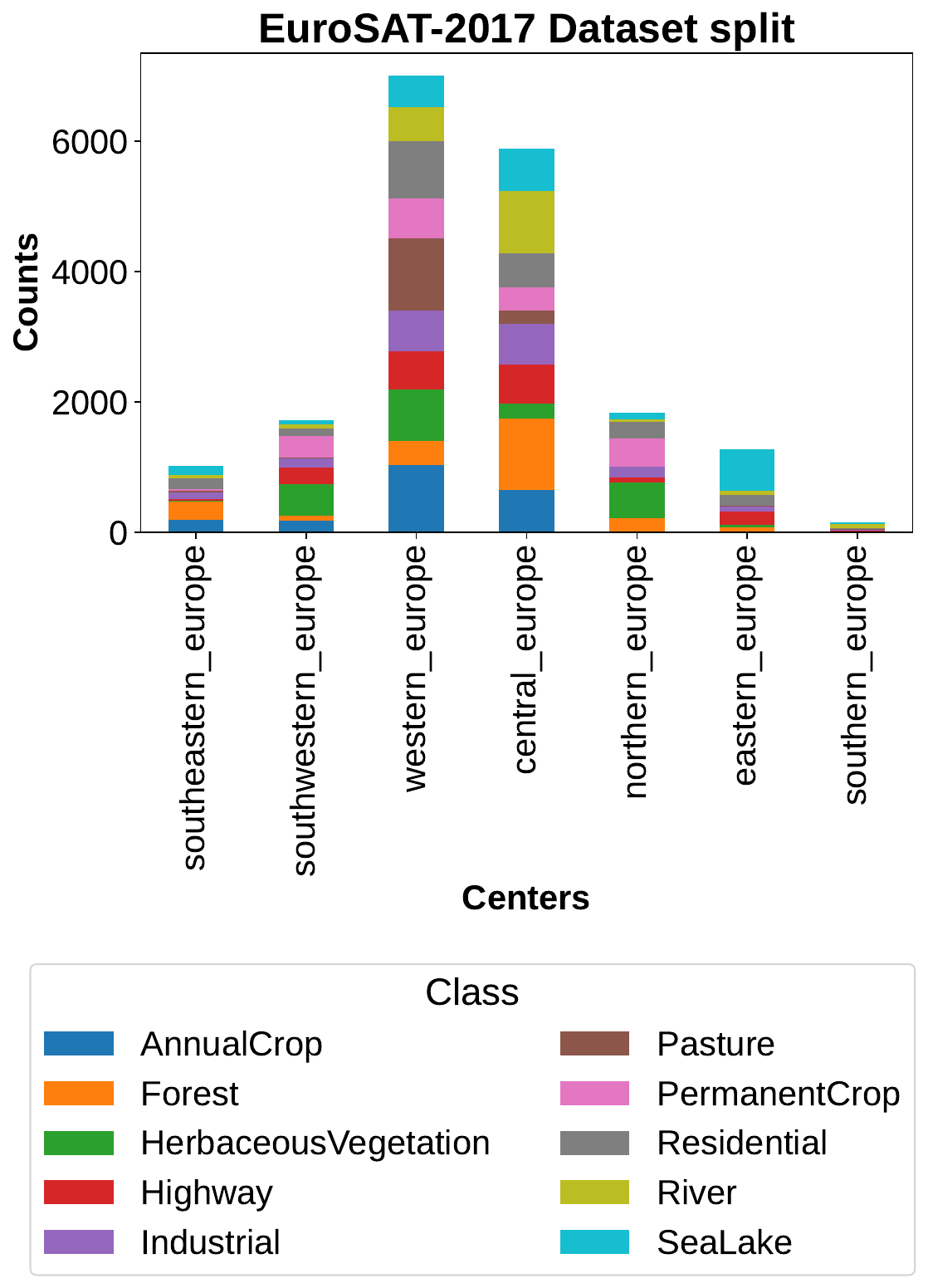}
\end{center}
   \caption{We split EuroSAT\cite{data-helber2019eurosat} into 7 clients based on subregion grouping according to ``The World Factbook". We use the geolocation information in GeoTIFF files to identify the countries of origin, based on which we assign images to specific clients.}
\label{fig:data-eurosat}
\end{figure}

\newpage

\begin{figure}[h]
\begin{center}
\includegraphics[width=0.75\linewidth]{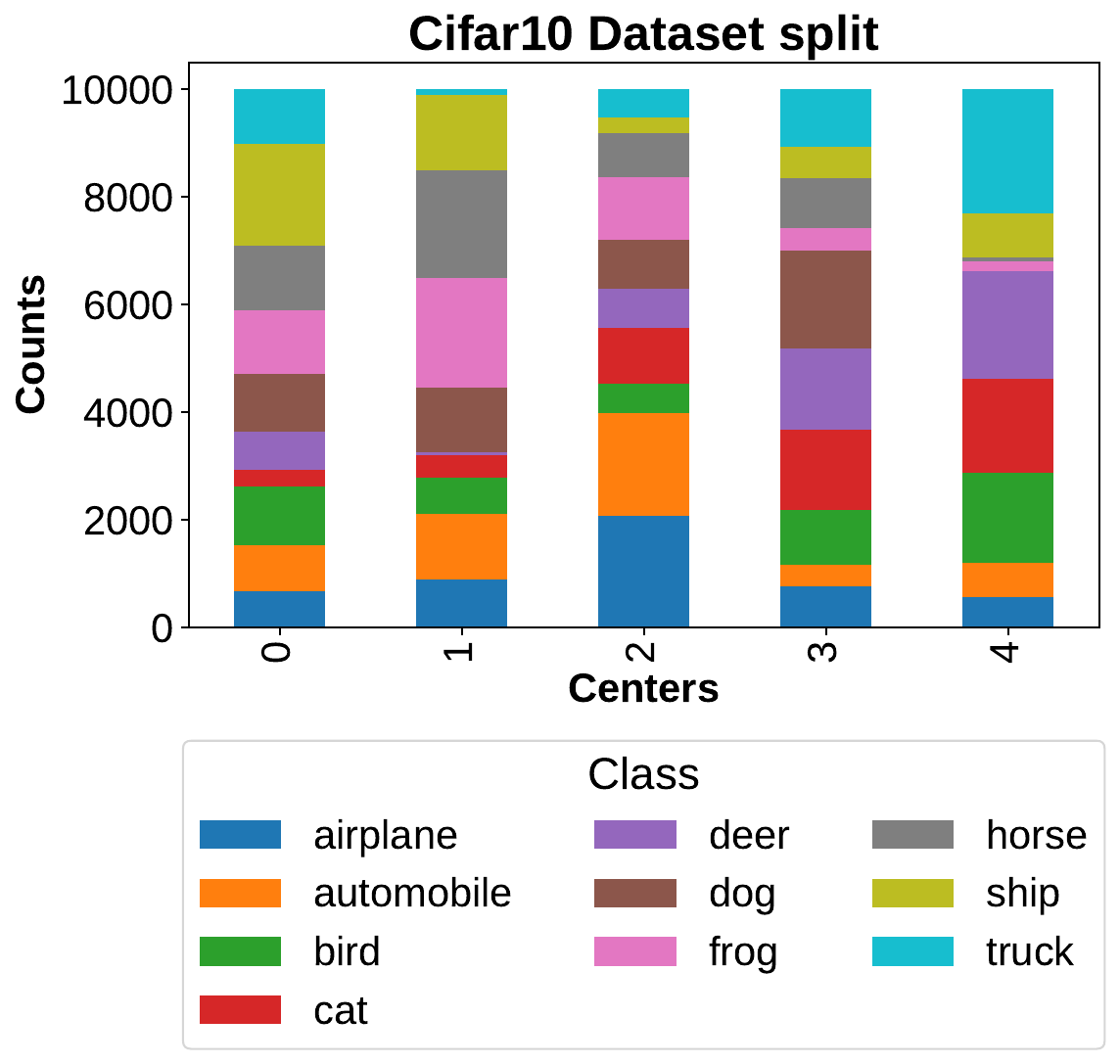}
\end{center}
   \caption{We split dataset into 5 clients with class imbalance. }
\label{fig:data-cifar10}
\end{figure}

\begin{figure}[h]
\begin{center}
\includegraphics[width=0.75\linewidth]{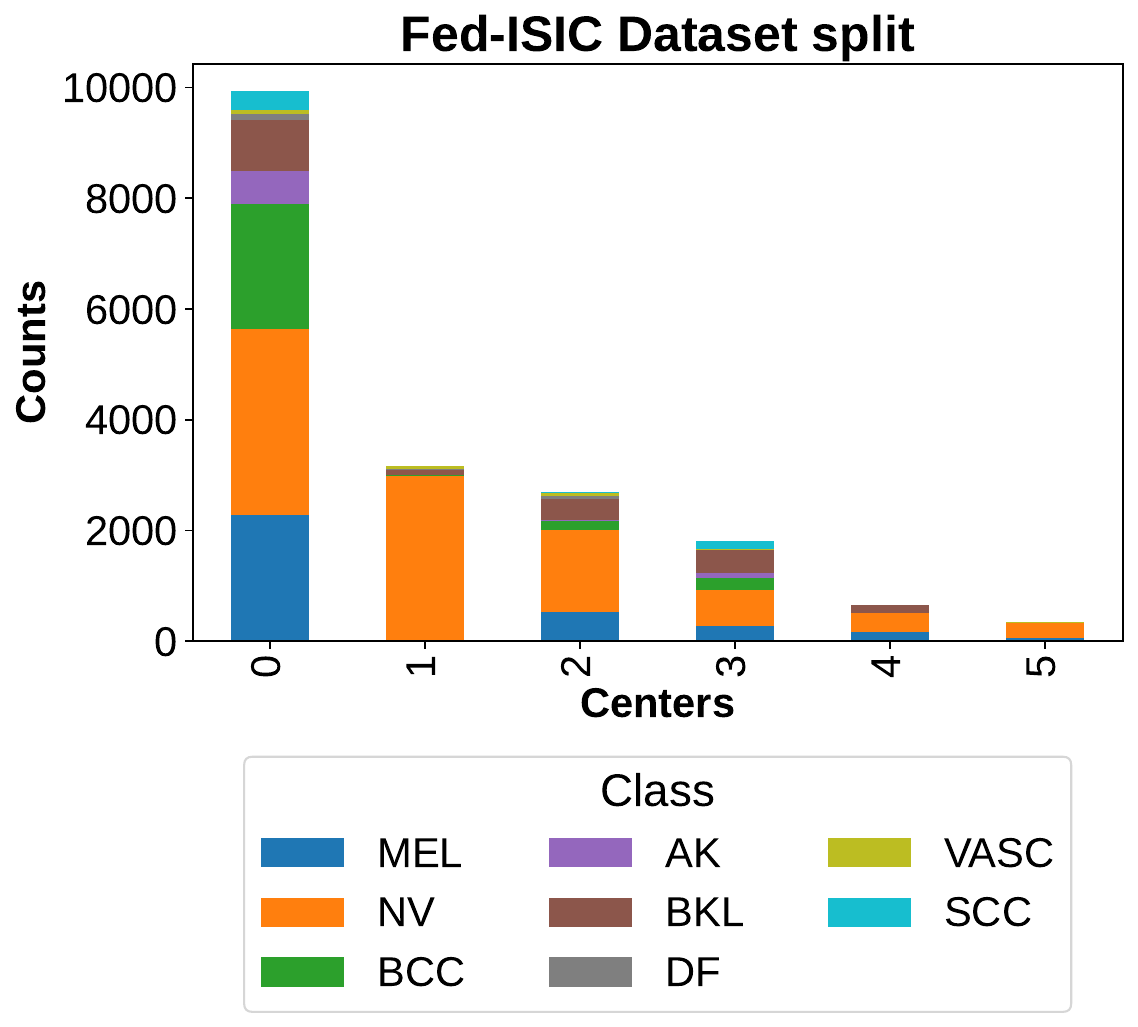}
\end{center}
   \caption{We use exact clients split in FedISIC\cite{data-ogier2022flamby}, which splits dataset based on devices and hospital of origin. }
\label{fig:data-isic}
\end{figure}

\newpage
\section{Training Setup}
For all the experiments, we use  F1-Score as a performance metric to account for class imbalances in test data. 
All the results are reported on a common pooled testset for the respective datasets. 
Collaboration happens at the end of every local training epoch wherein the clients would send all the trained models to the server.
Global aggregation in the server (\textaggfunc{RAggr}) is carried out once all models are received, and after the aggregation, a single model is broadcast to all the clients. This will be used as starting weights for the next epoch locally.
We call one single cycle of collaboration a round with a total of \textcount{T} rounds.
Since we take a cross-silo setting, we assume all clients will be available for all rounds throughout collaboration and assume hardware capabilities are homogeneous to train for a given task with a given model. 

\noindent\paragraph{FedSECA hyperparameters:} Our proposed \textaggfunc{RAggr} has 2 hyperparameters corresponding to sparsification and server momentum, which are kept the same for all experiments, irrespective of the dataset. 
We use the Sparsification factor $\gamma=0.9$, following it as a general rule of thumb from gradient compression and sparsification literature.
We set aggregator momentum $\beta_\texttt{ra}=0.5$

\noindent\paragraph{Local Training Setup:}
Local (clients) training hyperparameters are constant across all experiments, defense/attack settings, and ablations unless specified explicitly. 
We use AdamW optimizer with weight decay$=1\times 10^{-6}$ and betas of $\beta_1=0.9, \beta_2=0.999$ and optimizer state is reset for every epoch. For Cifar10 $lr=1\times 10^{-3}$ ran for $50$ epochs with a batch size of $64$ and image size $224$ and Kaiming weight initialization is used; for ISIC $lr=1\times 10^{-4}$ ran for $100$ epochs with a batch size of $32$ and image size of $200$ and Imagenet-1K pretrained weights is used as initialization; for EuroSAT $lr=1\times 10^{-5}$ ran for $50$ epochs with a batch size of $32$ and image size of $224$ and Imagenet-1K pretrained weights is used as initialization.

\noindent\paragraph{Comparison with Baselines:}
For experiments in \paperref{Paper-Fig.3}, the datasets are divided into cross-silo clients as shown in \cref{sup-sec:data_dist}. The Mean values of the last 5 communication rounds are tabulated in \cref{tab:fedrise_all_results} highlighting the trainings that collapsed or were severely impacted.
We use F1-Score as a performance metric considering class imbalances in test data. 
Experiments were run on NVIDIA GeForce RTX 4090 24GB machines.
For all three datasets, 2 clients in $4^{th}$ and $5^{th}$ rank are Byzantines.
We compare several robust aggregators with our method, and the exact formulations and hyperparameters of the compared robust aggregators are discussed in \cref{sec:defenses} and attacks are discussed \cref{sec:attacks}.

\noindent\paragraph{FedSECA Component Ablation:}
In the experiments reported in \paperref{Paper-Tab.1}, we disable and enable certain components in order to check the criticality of components in defending against specific attacks. We use the CIFAR10 split shown in \cref{fig:data-cifar10}.
Experiments were run on NVIDIA GeForce RTX 4090 24GB machines. The \paperref{Paper-Tab.1} second column also shows the variable that we set to enable or disable certain components.

\noindent\paragraph{FedSECA Hyperparameter Sensitivity:}
In the experiments in \paperref{Paper-Fig.4}, we study the effect of varying the two hyperparameters in the FedSECA. 
We use the CIFAR10 split shown in \cref{fig:data-cifar10}.
Experiments were run on NVIDIA GeForce RTX 4090 24GB machines.
For server momentum ($\beta_\texttt{ra}$), we experiment with the values $\{0.9,\;0.75,\;0.5,\;0.25,\;0.1,\;0.0\}$. And for the Sparsificaton ratio ($\gamma$), we experiment with values $\{0.99,\;0.9,\;0.82,\;0.67,\;0.51,\;0.22\}$ with 50 rounds collaboration.

\noindent\paragraph{Proportion of Byzantines:}
For experiments in \paperref{Paper-Fig.5}, we use CIFAR10-IID split with 64 clients where class label distribution is homogeneous. The batch size of $16$ is used for local training. The Byzantines counts are varied as $8(12\%),\; 17(25\%),\; 31(<50\%),\; 34(>50\%),\; 40(62\%)$ out of 64 clients. The experiment was run on  NVIDIA A100 SXM4 40GB machines.

\noindent\paragraph{Increasing Number of Clients:}
For experiments in \paperref{Paper-Tab.2}, we use CIFAR10-IID split with varying number of clients as $5,\; 10,\; 25,\; 50,\; 100,\; 200$. The batch size of $64$ and all local training hyperparameters are the same. The experiments were run on NVIDIA RTX A6000 49GB machines.
Additionally, the time taken by each robust aggregator function with varying numbers of clients is benchmarked in \cref{tab:all_raggr_timing} using CPU computation alone on Intel(R) Xeon(R) Silver 4215 CPU @ 2.50GHz. The FedSECA is designed for cross-silo applications where only a dozen clients are expected in collaboration; for the sake of being exhaustive, we compare the time for client counts from 4 to 256.

\cleardoublepage
\onecolumn 

\section{FedSECA}

\begin{table*}[h]
\centering
\resizebox{\linewidth}{!}{
\setlength{\tabcolsep}{4pt}
\begin{tabular}{ l| cccccccc cc| cccccccc cc| cccccccc c}
\toprule
\multirow{6}{*}{Robust Aggregator}
 & \multicolumn{9}{c}{\textbf{CIFAR10}} && \multicolumn{9}{c}{\textbf{EuroSAT}} && \multicolumn{9}{c}{\textbf{FedISIC}} \\ 
\midrule
 & \rotatebox{90}{ALIE} & \rotatebox{90}{IPM} & \rotatebox{90}{Fang} & \rotatebox{90}{LabelFlip} & \rotatebox{90}{Mimic} & \rotatebox{90}{Scale} & \rotatebox{90}{MinMax} & \rotatebox{90}{No Attack} & \rotatebox{90}{\textcolor{darkgray}{\textit{Mean}}}
 && \rotatebox{90}{ALIE} & \rotatebox{90}{IPM} & \rotatebox{90}{Fang} & \rotatebox{90}{LabelFlip} & \rotatebox{90}{Mimic} & \rotatebox{90}{Scale} & \rotatebox{90}{MinMax} & \rotatebox{90}{No Attack} & \rotatebox{90}{\textcolor{darkgray}{\textit{Mean}}}
 && \rotatebox{90}{ALIE} & \rotatebox{90}{IPM} & \rotatebox{90}{Fang} & \rotatebox{90}{LabelFlip} & \rotatebox{90}{Mimic} & \rotatebox{90}{Scale} & \rotatebox{90}{MinMax} & \rotatebox{90}{No Attack} & \rotatebox{90}{\textcolor{darkgray}{\textit{Mean}}}\\

\midrule

FedAvg (No Defense)      & \cellcolor{Tred}{0.10} & \cellcolor{Tred}{0.02} & \cellcolor{Tred}{0.02} & \cellcolor{Tyellow}{0.69} & 0.82 & \cellcolor{Tred}{0.02 } & \cellcolor{Tred}{0.02} & 0.88   & \textcolor{darkgray}{\textit{0.32}}
&&  0.97 & \cellcolor{Tyellow}{0.90} & \cellcolor{Tred}{0.02} & \cellcolor{Tyellow}{0.89} & 0.97 & \cellcolor{Tred}{0.02} &  \cellcolor{Tred}{0.02} & 0.98                                         & \textcolor{darkgray}{\textit{0.60}}
&&  0.71 & \cellcolor{Tyellow}{0.53} & \cellcolor{Tred}{0.01} & \cellcolor{Tyellow}{0.59} & 0.77 & \cellcolor{Tred}{0.04} &  \cellcolor{Tred}{0.08} & 0.77                                         & \textcolor{darkgray}{\textit{0.44}} \\

Krum            & \cellcolor{Tred}{0.02} & \cellcolor{Tred}{0.02} & \cellcolor{Tred}{0.04} & 0.76 & \cellcolor{Tyellow}{0.62} & 0.76 & \cellcolor{Tred}{0.02} & 0.76                                      & \textcolor{darkgray}{\textit{0.38}}
&&  0.98 & \cellcolor{Torange}{0.41} & \cellcolor{Torange}{0.54} & \cellcolor{Torange}{0.54} & 0.93 & \cellcolor{Torange}{0.54}& \cellcolor{Torange}{0.54} & \cellcolor{Torange}{0.54}                       & \textcolor{darkgray}{\textit{0.63}}
&&  \cellcolor{Torange}{0.24} & \cellcolor{Tred}{0.06} & \cellcolor{Torange}{0.22} & \cellcolor{Tred}{0.09} & 0.67 & \cellcolor{Torange}{0.22} & \cellcolor{Tred}{0.18} & \cellcolor{Torange}{0.22}       & \textcolor{darkgray}{\textit{0.24}} \\

RFA             & \cellcolor{Tred}{0.14} & 0\cellcolor{Tyellow}{.50} & \cellcolor{Tred}{0.02} & 0.80 & 0.71 & \cellcolor{Tyellow}{0.65} & \cellcolor{Tred}{0.02} & 0.88                   & \textcolor{darkgray}{\textit{0.46}}
&&  0.97 & \cellcolor{Tyellow}{0.70} & \cellcolor{Tred}{0.01} & 0.93 & 0.96 & \cellcolor{Tred}{0.14} &  \cellcolor{Tred}{0.01} & 0.97                                                     & \textcolor{darkgray}{\textit{0.59}}
&&  0.62 & \cellcolor{Torange}{0.29} & \cellcolor{Tred}{0.08} & \cellcolor{Tyellow}{0.51} & \cellcolor{Tyellow}{0.51} & \cellcolor{Tred}{0.04} & \cellcolor{Tred}{0.08} & 0.67            & \textcolor{darkgray}{\textit{0.35}} \\

CWTM            & \cellcolor{Tred}{0.04} & \cellcolor{Tred}{0.02} & \cellcolor{Tred}{0.05} & \cellcolor{Tyellow}{0.66} & 0.79 & \cellcolor{Tred}{0.02} & \cellcolor{Tred}{0.02} & 0.89    & \textcolor{darkgray}{\textit{0.31}}
&&  0.97 & \cellcolor{Tyellow}{0.58} & \cellcolor{Tred}{0.02} & \cellcolor{Tyellow}{0.88} & 0.96 & \cellcolor{Tred}{0.02} & \cellcolor{Tred}{0.02} & 0.97                                 & \textcolor{darkgray}{\textit{0.55}}
&&  \cellcolor{Tyellow}{0.61} & \cellcolor{Torange}{0.31} & \cellcolor{Tred}{0.08} & \cellcolor{Tyellow}{0.54} & 0.62 & \cellcolor{Tred}{0.04} & \cellcolor{Tred}{0.00} & 0.72            & \textcolor{darkgray}{\textit{0.36}} \\

HuberLoss & \cellcolor{Tred}{0.14}	&	0.72	&	\cellcolor{Tred}{0.04}	&	0.85	&	\cellcolor{Tyellow}{0.62}	&	0.77	&	\cellcolor{Tred}{0.02}	&	0.89	& \textcolor{darkgray}{\underline{0.51}}
&& 0.97	&	\cellcolor{Tyellow}{0.69}	&	0.78	&	0.93	&	0.95	&	0.96	&	0.78	&	0.97                & \textcolor{darkgray}{\textit{0.88}}
&& 0.61	&	\cellcolor{Torange}{0.29}	&	\cellcolor{Torange}{0.35}	&	\cellcolor{Tyellow}{0.50}	&	0.66	&	0.66	&	\cellcolor{Torange}{0.36}	&	0.67	            & \textcolor{darkgray}{\underline{0.51}} \\

CClipping       & \cellcolor{Torange}{0.46} & \cellcolor{Tyellow}{0.62} & \cellcolor{Tred}{0.02} & 0.85 & 0.83 & \cellcolor{Tred}{0.02} & \cellcolor{Tred}{0.02} & 0.88                   & \textcolor{darkgray}{\textit{0.46}}
&&  0.97 & \cellcolor{Tyellow}{0.90} & \cellcolor{Tred}{0.02} & \cellcolor{Tyellow}{0.89} & 0.97 & \cellcolor{Tred}{0.19} & \cellcolor{Tred}{0.02} & 0.98                                 & \textcolor{darkgray}{\textit{0.62}}
&&  0.71 & \cellcolor{Tyellow}{0.53 }& \cellcolor{Tred}{0.08} & \cellcolor{Tyellow}{0.59} & 0.76 & \cellcolor{Torange}{0.27} & \cellcolor{Tred}{0.00} & 0.77                              & \textcolor{darkgray}{\textit{0.46}} \\

CC-RandBucket   & \cellcolor{Tred}{0.06} & \cellcolor{Tred}{0.02} & \cellcolor{Tred}{0.02} & 0.72 & 0.83 & \cellcolor{Tred}{0.02} & \cellcolor{Tred}{0.02} & 0.88                         & \textcolor{darkgray}{\textit{0.32}}
&&  0.97 & \cellcolor{Tyellow}{0.88} & \cellcolor{Tred}{0.02} & \cellcolor{Tyellow}{0.79} & 0.97 & \cellcolor{Tred}{0.15} &  \cellcolor{Tred}{0.02} & 0.98                                & \textcolor{darkgray}{\textit{0.60}}
&&  0.71 & \cellcolor{Tyellow}{0.53} & \cellcolor{Tred}{0.01} & \cellcolor{Tyellow}{0.58} & 0.78 & \cellcolor{Tred}{0.16} &  0.08 & 0.76                                                     & \textcolor{darkgray}{\textit{0.45}} \\

CC-SeqBucket    & 0.79 & \cellcolor{Tyellow}{0.54} & \cellcolor{Tred}{0.02} & 0.84 & 0.80 & \cellcolor{Tred}{0.02} & \cellcolor{Tred}{0.02} & 0.88                                        & \textcolor{darkgray}{\textit{0.49}}
&&  0.95 & 0.88 & \cellcolor{Tred}{0.02} & \cellcolor{Torange}{0.50} & 0.96 & \cellcolor{Tred}{0.15} & \cellcolor{Tred}{0.02} & 0.96                                                      & \textcolor{darkgray}{\textit{0.55}}
&&  0.71 & \cellcolor{Tyellow}{0.47} & \cellcolor{Tred}{0.02} & \cellcolor{Tyellow}{0.45} & 0.73 & \cellcolor{Torange}{0.24}  &  \cellcolor{Tred}{0.00} & 0.72                            & \textcolor{darkgray}{0.42} \\

COPOD-DOS       & 0.82 & \cellcolor{Torange}{0.47} & \cellcolor{Tred}{0.04} & 0.84 & \cellcolor{Tyellow}{0.63} & \cellcolor{Tred}{0.02} & \cellcolor{Tred}{0.03} & 0.87                                                       & \textcolor{darkgray}{\textit{0.46}}
&&  0.95 & \cellcolor{Torange}{0.22} & 0.93 & 0.93 & 0.92 & 0.92 &  \cellcolor{Tred}{0.05} & 0.93                                                                                                                             & \textcolor{darkgray}{\textit{0.73}}
&&  \cellcolor{Tyellow}{0.42} & \cellcolor{Tred}{0.13} & \cellcolor{Tyellow}{0.51} & \cellcolor{Tyellow}{0.27} & \cellcolor{Tyellow}{0.42} & \cellcolor{Tred}{0.10}  &  \cellcolor{Tred}{0.08} & \cellcolor{Tyellow}{0.49}    & \textcolor{darkgray}{\textit{0.30}} \\

FL-Detector & \cellcolor{Tred}{0.02}	&	\cellcolor{Tyellow}{0.52}	&	\cellcolor{Torange}{0.46}	&	\cellcolor{Tred}{0.06}	&	\cellcolor{Torange}{0.44}	&	0.81	&	\cellcolor{Tyellow}{0.70}	&	0.88  & \textcolor{darkgray}{\textit{0.49}}
&& 0.96	&	\cellcolor{Tyellow}{0.72}	&	0.93	&	0.93	&	0.97	&	0.94	&	0.93	&	0.97              & \textcolor{darkgray}{\textit{\underline{0.92}}}
&& \cellcolor{Torange}{0.37}	&	\cellcolor{Torange}{0.24}	&	\cellcolor{Tyellow}{0.57}	&	\cellcolor{Tyellow}{0.50}	&	\cellcolor{Tyellow}{0.44}	&	\cellcolor{Tyellow}{0.57}	&	\cellcolor{Tyellow}{0.57}	&	0.66	           & \textcolor{darkgray}{\textit{0.49}} \\           

TIES-Merging    & 0.85 & \cellcolor{Tred}{0.02} & 0.73 & 0.72 & 0.81 & \cellcolor{Tred}{0.02}  & \cellcolor{Tred}{0.03} & 0.82                                                    & \textcolor{darkgray}{\textit{0.5}}
&&  0.98 & 0.92 & \cellcolor{Tred}{0.02} & 0.91 & 0.97 & \cellcolor{Tred}{0.02} &  \cellcolor{Tred}{0.01} & 0.98                                                                  & \textcolor{darkgray}{\textit{0.60}}
&&  0.73 & \cellcolor{Tred}{0.00}& \cellcolor{Tred}{0.05} & \cellcolor{Tyellow}{0.60} & 0.72 & \cellcolor{Tred}{0.04} & \cellcolor{Tred}{0.00} & 0.75                             & \textcolor{darkgray}{\textit{0.36}} \\

FedSECA (Ours)   & 0.72 & 0.82 & 0.78 & 0.76 & 0.77 & 0.72 & 0.83 & 0.81                                                                                                                        & \textcolor{darkgray}{\textbf{0.78}}
&&  0.96 & 0.95 & 0.94 & 0.94 & 0.96 & 0.94 & 0.94 & 0.97                                                                                                                                       & \textcolor{darkgray}{\textbf{0.95}}
&&  0.64 & \cellcolor{Tyellow}{0.46} & \cellcolor{Tyellow}{0.47} & \cellcolor{Tyellow}{0.52} & \cellcolor{Tyellow}{0.56} & \cellcolor{Tyellow}{0.53} & \cellcolor{Tyellow}{0.45} & \cellcolor{Tyellow}{0.51}         & \textcolor{darkgray}{\textbf{0.52}} \\

\midrule
\bottomrule
\end{tabular}
}
  \caption{
  \textbf{\textit{Main Results as Table:}} This corresponds to the main result in \paperref{Paper-Fig.3} showing convergence of various \textaggfunc{RAggr} uder attacks. Values shown are the mean F1-score of the last 5 comm-rounds computed on the test set. It can be clearly seen that all defenses collapse, at least for some of the attacks, whereas FedSECA is robust to all attacks. Only in the FedISIC dataset, the overall results are suboptimal, yet it does not collapse for any of the attacks, managing to converge. \\ 
    \colorbox{Tred}{\footnotesize Red} indicates that the training collapsed due to the attack, cutoff at $0.2$ for all three datasets.  \\
    \colorbox{Torange}{\footnotesize Orange} indicates that the training was impacted severely, cutoff at $0.5$(CIFAR10), $0.55$(EuroSAT) and $0.4$(FedISIC). \\
    \colorbox{Tyellow}{\footnotesize Yellow} indicates a noticeable drop in accuracy, cutoff at $0.7$(CIFAR10), $0.9$(EuroSAT) and $0.6$(FedISIC). \\
    }
  \label{tab:fedrise_components
  }
  \label{tab:fedrise_all_results}
\end{table*}

\begin{table*}[h]
\centering
\resizebox{0.75\linewidth}{!}{
\setlength{\tabcolsep}{4pt}
\begin{tabular}{ lc| rrrrrrr }
\toprule
\midrule
Client Count &&  4  &  8  &  16  &  32  &  64  &  128  & 256 \\
\midrule
FedAvg          &&  {0.20}{\scriptsize$\pm$0.01}  &  {0.35} {\scriptsize$\pm$0.01}   &  {0.50 }{\scriptsize$\pm$0.05}  &  {1.37  }{\scriptsize$\pm$0.02}  &  {2.69  }{\scriptsize$\pm$0.14}  &  {5.69   }{\scriptsize$\pm$0.26}  &  {11.52  }{\scriptsize$\pm$0.51}   \\
Krum            &&  {0.59}{\scriptsize$\pm$0.02}  &  {1.79} {\scriptsize$\pm$0.01}   &  {6.28 }{\scriptsize$\pm$0.01}  &  {24.49 }{\scriptsize$\pm$0.26}  &  {95.84 }{\scriptsize$\pm$0.39}  &  {383.56 }{\scriptsize$\pm$0.35}  &  {1606.68}{\scriptsize$\pm$3.75}   \\
RFA             &&  {0.86}{\scriptsize$\pm$0.02}  &  {1.48} {\scriptsize$\pm$0.01}   &  {2.61 }{\scriptsize$\pm$0.02}  &  {5.69  }{\scriptsize$\pm$0.04}  &  {11.54 }{\scriptsize$\pm$0.21}  &  {23.80  }{\scriptsize$\pm$0.40}  &  {48.55  }{\scriptsize$\pm$0.26}   \\
CWTM            &&  {1.62}{\scriptsize$\pm$0.02}  &  {3.46} {\scriptsize$\pm$0.01}   &  {8.20 }{\scriptsize$\pm$0.06}  &  {19.76 }{\scriptsize$\pm$0.05}  &  {44.42 }{\scriptsize$\pm$0.47}  &  {99.80  }{\scriptsize$\pm$0.92}  &  {219.33 }{\scriptsize$\pm$0.71}   \\
HuberLoss       &&  {1.47}{\scriptsize$\pm$0.01}  &  {2.04} {\scriptsize$\pm$0.01}   &  {2.73 }{\scriptsize$\pm$0.01}  &  {5.81  }{\scriptsize$\pm$0.01}  &  {11.50 }{\scriptsize$\pm$0.21}  &  {23.60  }{\scriptsize$\pm$0.01}  &  {35.64  }{\scriptsize$\pm$0.46}   \\
CClipping       &&  {1.21}{\scriptsize$\pm$0.01}  &  {2.20} {\scriptsize$\pm$0.01}   &  {4.04 }{\scriptsize$\pm$0.00}  &  {8.49  }{\scriptsize$\pm$0.04}  &  {16.92 }{\scriptsize$\pm$0.22}  &  {35.09  }{\scriptsize$\pm$0.52}  &  {70.87  }{\scriptsize$\pm$0.41}   \\
CC-RandBucket   &&  {0.97}{\scriptsize$\pm$0.01}  &  {1.77} {\scriptsize$\pm$0.02}   &  {2.63 }{\scriptsize$\pm$0.02}  &  {6.51  }{\scriptsize$\pm$0.03}  &  {12.58 }{\scriptsize$\pm$0.57}  &  {26.87  }{\scriptsize$\pm$0.22}  &  {53.06  }{\scriptsize$\pm$0.13}   \\
CC-SeqBucket    &&  {1.11}{\scriptsize$\pm$0.01}  &  {2.14} {\scriptsize$\pm$0.02}   &  {2.91 }{\scriptsize$\pm$0.13}  &  {8.39  }{\scriptsize$\pm$0.06}  &  {16.97 }{\scriptsize$\pm$0.20}  &  {34.84  }{\scriptsize$\pm$0.11}  &  {70.39  }{\scriptsize$\pm$0.08}   \\
COPOD-DOS       &&  {2.66}{\scriptsize$\pm$0.17}  &  {8.70} {\scriptsize$\pm$0.05}   &  {32.68}{\scriptsize$\pm$0.12}  &  {127.52}{\scriptsize$\pm$0.03}  &  {503.49}{\scriptsize$\pm$0.06}  &  {2036.33}{\scriptsize$\pm$4.08}  &  {8663.10}{\scriptsize$\pm$24.31}  \\
FL-Detector     &&  {1.55}{\scriptsize$\pm$0.37}  &  {2.81} {\scriptsize$\pm$0.56}   &  {5.89 }{\scriptsize$\pm$1.06}  &  {12.51 }{\scriptsize$\pm$2.58}  &  {23.49 }{\scriptsize$\pm$4.88}  &  {46.40  }{\scriptsize$\pm$9.26}  &  {94.14  }{\scriptsize$\pm$19.73}  \\
TIES-Merging    &&  {1.89}{\scriptsize$\pm$0.04}  &  {3.36} {\scriptsize$\pm$0.07}   &  {6.25 }{\scriptsize$\pm$0.10}  &  {12.40 }{\scriptsize$\pm$0.21}  &  {24.71 }{\scriptsize$\pm$0.89}  &  {51.46  }{\scriptsize$\pm$1.65}  &  {107.06 }{\scriptsize$\pm$1.40}   \\
FedSECA (Ours)  &&  {4.87}{\scriptsize$\pm$0.00}  &  {12.30}{\scriptsize$\pm$0.04}   &  {37.79}{\scriptsize$\pm$0.18}  &  {125.27}{\scriptsize$\pm$0.24}  &  {456.32}{\scriptsize$\pm$0.45}  &  {1791.72}{\scriptsize$\pm$3.72}  &  {8073.35}{\scriptsize$\pm$25.93}  \\

\midrule
\bottomrule
\end{tabular}
}
  \caption{
  \textbf{\textit{Time for single aggregation call of each \textaggfunc{RAggr}}}: The times shown are in seconds it took to complete aggregation operation. We report each aggregator function's mean and standard deviation by repeating the function 5 times. We use the ResNet18 model with a parameter size of 11.2M.
  Benchmarking is conducted with a CPU with 8 virtual cores node (AMD EPYC 9654) without GPU.
  We implement all the aggregator functions from scratch using torch tensors and numpy arrays using standard library functions without using any approximations or operation-specific optimizations.
  It can be observed that FedSECA is suitable for Cross-Silo settings, and it takes very long with a large number of clients but provides superior robustness to various attacks as shown in \paperref{Paper-Fig.3}. For Cross-Device applications, we recommend a client selection approach before robust aggregation.
  }
  \label{tab:fedrise_components
  }
  \label{tab:all_raggr_timing}
\end{table*}


\lstset{
    language=Python,
    basicstyle=\small\ttfamily\bfseries,
    keywordstyle=\color{magenta},
    commentstyle=\color{teal},
    stringstyle=\color{pink},
    numbers=left,
    numberstyle=\scriptsize\sffamily\color{gray},
    stepnumber=1,
    numbersep=10pt,
    backgroundcolor=\color{white},
    showspaces=false,
    showstringspaces=false,
    showtabs=false,
    tabsize=4,
    captionpos=b,
    breaklines=true,
    breakatwhitespace=false,
    escapeinside={\%*}{*)}
}


\begin{algorithm*}[!htb]
  \caption{\textbf{\textaggfunc{FedSECA}} Gradient Aggregation \textemdash\: \textit{PyTorch pseudocode}}
\begin{lstlisting}
## Initialize
K     = number_of_clients
beta  = 0.5
gamma = 0.9
g     = torch.vstack([g1, g2, g3,...,gk])  # all clients gradients
g_tminus1 = g_sent_in_previously           # will be initialized to 0

def FedSECA_aggregator(g, g_tminus1, gamma, beta, K):
    ## Magnitude of Gradients
    magn_g = torch.abs(g)
    
    ## Sign of Gradients
    sign_g = torch.sign(g)
    
    ###### CRISE block =========================
    
    ## Concordance Ratio Computation
    rho_list = []
    for i in range(K):
        omega_c = torch.F.cosine_similarity(sign_g.view(K,-1) , sign_g[i].view(1,-1))
        rho_i = torch.sign(omega_c).mean()
        rho_list.append(rho_i)
    repute_rho = torch.vstack(rho_list).view(K,1)
    repute_rho = torch.clamp(repute_rho, min=0)    

    ###### RoCA block =========================
    
    ## Gradient Clipping
    g_norms = torch.norm(g, dim=1)
    med_norm, _ = torch.median(g_norms, dim=0)
    norm_clip = torch.minimum(torch.tensor(1.0), med_norm/g_norms)
    g_clipped = g.view(K,-1) * norm_clip.view(K, 1)
    
    ## Gradient Clamping
    med_magn, _ = torch.median(torch.abs(g_clipped), dim=0)
    g_hat = torch.clamp(g, max=med_magn, min=-med_magn)
            
    ## Top-K Sparsification
    q = magn_g.quantile(gamma, dim=1).view(-1, 1)
    g_ddot = g_hat; g_ddot[magn_g<q] = 0.0

    #####
    g_star = g_ddot
    
    ## Sign Voting
    sign_g_star = torch.sign(g_star)
    voted_sign = (sign_g_star*repute_rho).sum(dim=0).view(1,-1)
        
    ##  Coordinate-wise Gradient aggregation 
    g_mask     = (0<(g_star * voted_sign)).int()
    g_selected = g_star * g_mask
    g_divisor  = g_mask.sum(dim=0)
    g_aggregate = safe_divide(g_selected.sum(dim=0), g_divisor, fill=0.0)
    
    ## Server Momentum 
    g_send = (1-beta)*g_aggregate + beta*g_tminus1
    g_tminus1 = g_send
    
    return g_send
\end{lstlisting}
\end{algorithm*}

\cleardoublepage
\subsection{Concordance Ratio in CRISE}

\begin{figure*}[h]
    \includegraphics[width=\linewidth]{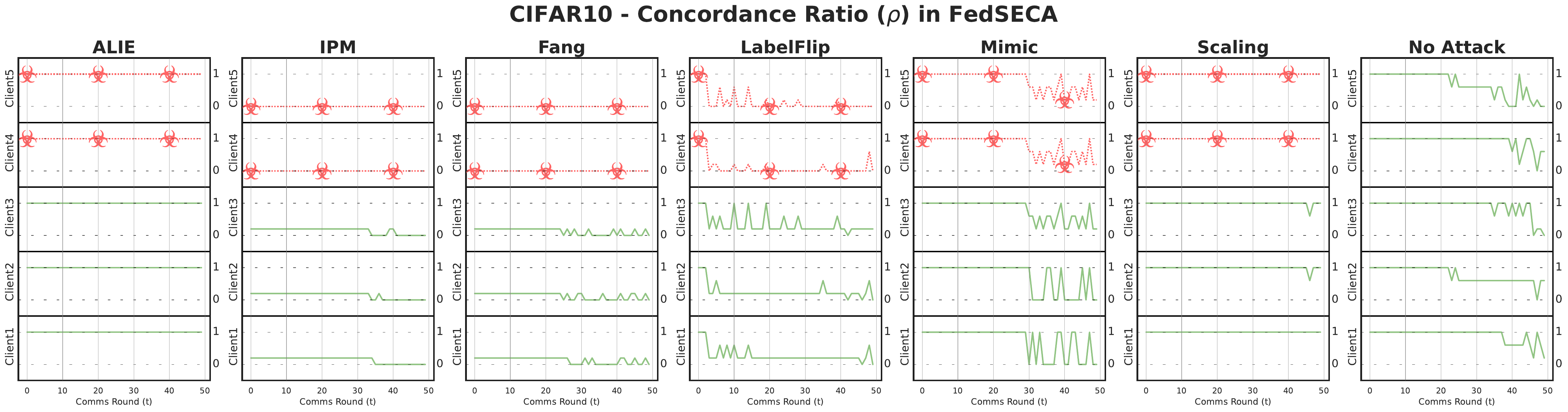}
    \includegraphics[width=\linewidth]{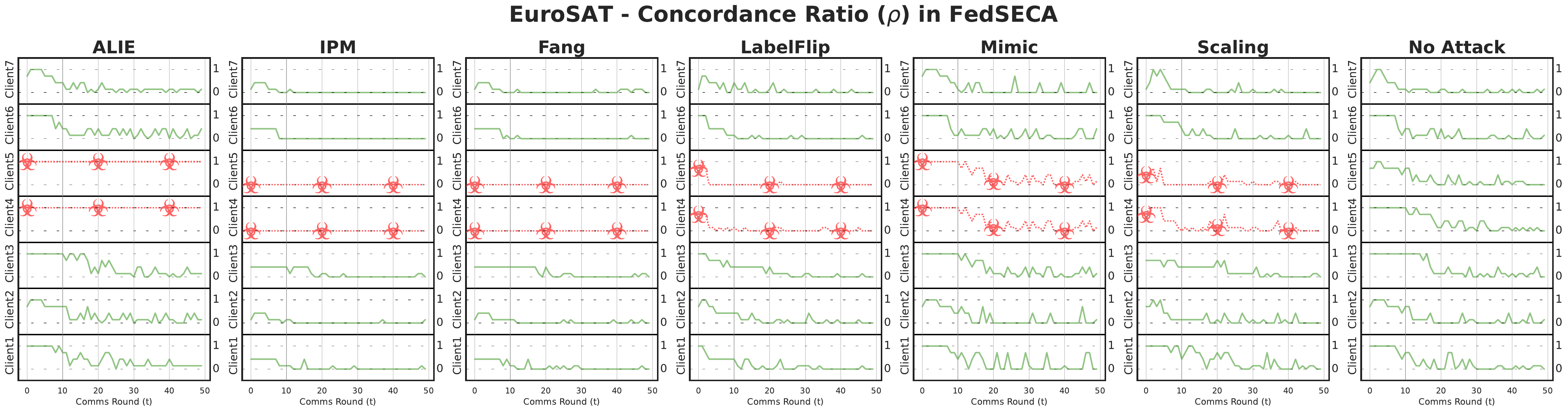}
    \includegraphics[width=\linewidth]{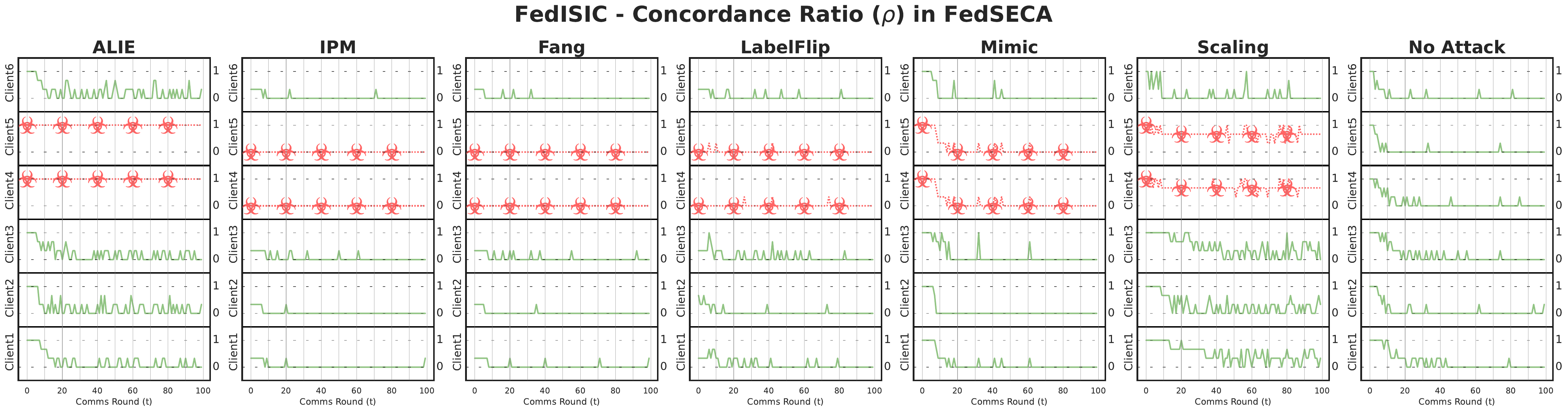}
    \caption{\textbf{Gradients Concordance Ratio ($\rho$):} Evolution of $\rho$ values for each client throughout the training computed in CRISE. \textcolor{Green}{Green} lines correspond to honest clients, \textcolor{Red}{Red} lines correspond to malicious clients.
    Client-level Concordance Ratio is used solely for voting weightage; once the optimal gradient signs are chosen for each parameter, the gradient filtering follows an egalitarian approach only considering the alignment of the signs but not the $\rho$.
    If we filter at the client level solely based on $\rho$, there is a high chance that an honest client might be ignored in its entirety and malicious clients might be included, especially as seen in the ALIE attack.
    Although ALIE evades the detection while computing $\rho$, it will be filtered out in the sparsification step, criticality of the sparsification step in neutralizing ALIE is also shown in \paperref{Paper-Tab.1}.
    Although it can be seen that some honest clients might have lower weightage, especially later in training, this has little effect on convergence.}
    \label{fig:rating-values}
\end{figure*}

\twocolumn 
\cleardoublepage
Let $\textset{K}$ represent the set of all clients involved in the federation, $\textcount{K} = |\textset{K}|$ total count of clients in the federation.
Let $\textset{B}$ represent a set of Byzantine or malicious clients present in the consortium, $\textcount{B} = |\textset{B}|$ total count of Byzantine clients.
Let $\textset{H}$ represent a set of Honest clients, $\textcount{H}=|\textset{H}| = (\textcount{K} - \textcount{B})$ total count of Honest clients.
Here $\textset{B} \subset \{1,2,..,k\}$  and $\textset{H} \subset \{1,2,..,k\}$ such that set of all clients is given as $\textset{K} = \textset{B}\uplus\textset{H}$ and total count is $|\textset{K}|=|\textset{H}|+|\textset{B}|$.
Let $\textcount{D}$ represent the number of parameters in the model $\textset{M}$.

\textit{The rest of the variable names (and Greek letters) henceforth defined are for specific attack methods and robust aggregators. These variables are different from the ones used in FedSECA in main paper unless specified otherwise.}
\section{Attacks Formulation} \label{sec:attacks}
$\textattack{\textvec{w}}$ or $\textattack{\textvec{g}}$ represents poisoned weights or gradients sent by byzantine clients.
Let $\delta$ be small noise added to the attack hyperparameters to make the attack vector slightly different from each other to evade any heuristic checks, here $\delta~\sim~\textstat{Uniform}(-0.05,0.05)$

\subsection{Crafting Parameter Direction (Fang)}
Fang attack\cite{attk-fang2020local} shifts the global model parameter in a manner that counters its actual direction of change in the absence of any attacks. $\textvec{s}$ indicates an increase or decrease (sign) of the global model parameter relative to the previous iteration. Thereby the objective is to perturb the weights in the opposite direction to $\textvec{s}$

\begin{equation}
\begin{aligned}
    \textattack{\textvec{w}}_b^{(t)} =&\; \widetilde{\textvec{w}}^{(t-1)} - \lambda_\texttt{f}\cdot \textvec{s}
    \qquad \forall\: b \in \textset{B}\\
    \textvec{s} =&\; \textstat{sgn} ( \mu_{\textvec{w}_{\textset{H}}}^{(t)} -  \widetilde{\textvec{w}}^{(t-1)}) \\
    \textnormal{where} \;&\; \mu_{\textvec{w}_{\textset{H}}}^{(t)} =\; \frac{1}{\textcount{H}} \sum_{k \in \textset{H}} \textvec{w}_{k}^{(t)}
\end{aligned}
\label{eq:attack-fang}
\end{equation}

Here $\lambda_\texttt{f}$ is the strength of the attack, which can be adjusted as required. Here we use $\lambda = 0.1 + \delta$. $\textstat{sgn}$ is the signum function.

\subsection{A Little is Enough (ALIE)}

In ALIE\cite{attk-baruch2019little}, the authors demonstrate that by consistently applying minor alterations to numerous parameters, a malicious client can disrupt the convergence of the global model. Given the normal distribution of data, if the attacker has access to a representative subset of the workers, only the corrupted workers' data is needed to estimate the mean and standard deviation of the distribution, thereby facilitating the manipulation of models.

\begin{equation}
\begin{aligned}
    \textattack{\textvec{g}}_b^{(t)} =& \mu_{\textvec{g}_{\textset{H}}}^{(t)} - z_\texttt{max} \cdot \sigma_{\textvec{g}_{\textset{H}}}^{(t)} 
    \qquad \forall\: b \in \textset{B}\\
    \textnormal{where} \quad&\quad \mu_{\textvec{g}_{\textset{H}}}^{(t)} = \frac{1}{\textcount{H}} \sum_{k \in \textset{H}} \textvec{g}_{k}^{(t)}  \quad, \\
    &\quad \sigma_{\textvec{g}_{\textset{H}}}^{(t)} = \sqrt{ \frac{1}{\textcount{H}} \sum_{k \in \textset{H}}(\textvec{g}_{k}^{(t)} - \mu_{\textvec{g}_{\textset{H}}}^{(t)})^2 }
\end{aligned}
\label{eq:attack-alie}
\end{equation}

Here, $z_\texttt{max}$ corresponds to the factor determining the standard deviation shift of the attack vector from actual honest updates. In literature this is computed based on the number of clients involved, which was ineffective under a cross-silo setting, thus we choose an empirical value based on the impact on no defense setting. We set $z_\texttt{max} = 1.0 + \delta$.

\subsection{Inner Product Manipulation (IPM)}

The basis for the IPM\cite{attk-xie2020fall} lies in the observation that as the gradient descent algorithm approaches convergence, the gradient $\textvec{g}$ tends toward 0. Consequently, despite the bounded distance between the robust estimator and the true mean, it remains feasible to manipulate their inner product to yield a negative value.

\begin{equation}
    \langle\: \nabla L(\textvec{w}), \textaggfunc{Aggr}( \{\textvec{g}_k : k \in \textset{K}\}) \: \rangle \geq 0
\label{eq:attk-ipm_converge}
\end{equation}

In gradient descent algorithms, convergence in the loss is guaranteed only if the inner product between the true gradient $\nabla L(\textvec{w})$ and the robust estimator remains non-negative as shown \ref{eq:attk-ipm_converge}. Hence the potential attack vector can be formed as below:

\begin{equation}
\begin{aligned}
    \textattack{\textvec{g}}_b^{(t)} =&\; - \epsilon \cdot \mu_{\textvec{g}_{\textset{H}}}^{(t)}
    \qquad \forall\: b \in \textset{B} \\
    \textnormal{where}&\quad \mu_{\textvec{g}_{\textset{H}}}^{(t)} = \frac{1}{\textcount{H}} \sum_{k \in \textset{H}} \textvec{g}_{k}^{(t)} \\
\end{aligned}
\label{eq:attack-ipm}
\end{equation}

Here in IPM, $\epsilon$ is the strength of the attack, which can be adjusted to increase or decrease attack intensity while remaining undetected. Here we use $\epsilon = 1.3 + \delta$.

\subsection{Mimic}
The objective of the Mimic attack proposed in \cite{defns-karimireddy2020byzantine_randBuck} is to maximize the data imbalance perceived by aggregators, especially middle-seeking approaches, even in balanced datasets. This works by overemphasizing a specific set of clients in margins by mimicking their updates, thereby forming a majority. Because of this, clients with rich gradients will be excluded. The severity of the attack will increase with the increase in heterogeneity among actual distribution and type of defense mechanism used.
For identifying the client to mimic $k^\star$, we compute the direction $\textvec{z}$ of maximum variance of the honest workers' gradients. The client to mimic during the warmup steps are determined as follows

\begin{equation}
\begin{aligned}
    & \qquad \textattack{\textvec{g}}_b^{(t+1)} =\; \textvec{g}_{k^\star}^{(t+1)}  \\
    {k^\star} =&\; \arg\!\max_{k \in \textset{H}}\; \bigl[{\textvec{z}^{(t+1)}}^\top  \textvec{g}^{(t+1)}_{k}\bigr] 
     \qquad \forall\: b \in \textset{B} \\
    \textvec{z}^{(t+1)} =&\; \arg\!\max_{\|\textvec{z}\|=1}\; 
    \left[ {\textvec{z}^{(t)}}^\top \Bigl( \sum_{t \in \textset{T}} \sum_{k \in \textset{H}} 
    (\textvec{g}_{k}^{(t)} - \mu_{\textvec{g}_{\textset{H}}}^{(t)}) \times \right. \\
    & \left. \qquad\qquad\qquad\qquad\qquad (\textvec{g}_{k}^{(t)} -
    \mu_{\textvec{g}_{\textset{H}}}^{(t)})^\top \Bigr) \textvec{z}^{(t)} \right] \\
    \text{where} &\quad \mu_{\textvec{g}_{\textset{H}}}^{(t)} = \frac{1}{|\textset{H}|\,|\textset{T}|} \sum_{k \in \textset{H}} \sum_{t \in \textset{T}} \textvec{g}_{k}^{(t)} \\    
\end{aligned}
\label{eq:attk-mimic}
\end{equation}

\noindent the computation of $\textvec{z}$ can be approximated as below,

\begin{equation}
\begin{aligned}
    \textvec{z}^{(t+1)} \approx&\; \frac{t}{1 + t} \textvec{z}^{(t)} + \\
     & \frac{1}{1 + t} \sum_{k \in \textset{H}} (\textvec{g}^{(t+1)}_k - \mu_{\textvec{g}_{\textset{H}}}^{(t+1)})(\textvec{g}^{(t+1)}_k - \mu_{\textvec{g}_{\textset{H}}}^{(t+1)})^\top \textvec{z}^{(t)} \\
    \text{where} &\quad \mu_{\textvec{g}_{\textset{H}}}^{(t+1)} = \frac{t}{1 + t} \mu_{\textvec{g}_{\textset{H}}}^{(t)} + \frac{1}{(1 + t) |\textset{H}|} \sum_{k \in \textset{H}} \textvec{g}_{k}^{(t+1)}\\    
\end{aligned}
\label{eq:attk-mimic_approx}
\end{equation}

After the warmup steps, the mimicked client is kept constant ${k^\star}$ obtained at $t$ for the rest of the training. $\textvec{z}$ is initialized randomly at the beginning of the training and recomputed at each step during the warmup steps.

\subsection{LabelFlip }

Suppose, $\textset{C}$ set classes in the classification task, label flipping involves inverting labels such that prediction $\hat{y}_i$ each image maps to the wrong class $y^\prime_i$ instead of true $y_i$.

\begin{equation*}
    y^\prime_i = \textcount{C} - y_i \quad \text{where} \quad y \in \textset{C} \quad \text{and} \quad \textcount{C}=|\textset{C}|
\label{eq:attk-labelflip}
\end{equation*}

such that the categorical cross-entropy optimization of a Byzantine client becomes 

\begin{equation}
    \textattack{\textvec{w}}_{b}^{(t)} = \widetilde{\textvec{w}}^{(t+1)} - \eta \cdot \frac{1}{\textcount{N}_k} \sum_{i=1}^{\textcount{N}_k} \textfunc{L}(\textvec{w}; x_i, y^\prime_i) 
    \qquad \forall\: b \in \textset{B}
\label{eq:attack-lflip}
\end{equation}

This would perturb the gradients, especially ones that are closer to the classification head, in the wrong direction; thus, the majority of the gradients will follow the actual direction, thereby remaining closer to other honest clients, while the critical gradients' specific task at hand will be poisoned.

\subsection{Scaling}

In Scaling attack, the original gradients are scaled by a factor $\varepsilon$, and this will derail the training.
If the scaling is positive, then the direction of gradients will align with the true gradients, but the value will be higher than the regular values. This will cause the gradients to explode and destabilize the training.
If the directions are flipped, this becomes equivalent to an IPM attack. In this work, we set the scaling factor as $\varepsilon=10$.

\begin{equation}
\begin{aligned}
    \textattack{\textvec{g}}_b^{(t)} =&\; \varepsilon \cdot \mu_{\textvec{g}_{\textset{H}}}^{(t)}
    \qquad \forall\: b \in \textset{B} \\
    \textnormal{where}&\quad \mu_{\textvec{g}_{\textset{H}}}^{(t)} = \frac{1}{\textcount{H}} \sum_{k \in \textset{H}} \textvec{g}_{k}^{(t)} \\
\end{aligned}
\label{eq:attack-scaling}
\end{equation}

\subsection{Adaptive Attack}
For optimization-based adaptive attacks, we use Min-Max\cite{attk-hejwalkar2021minmax_minsum_dnc} attack, with sign flipping (similar to \cite{attk-fang2020local}) where the attack strength is adaptively optimized for the defense employed based on the impact it can make on a given set of updates from honest clients. This version of the attack that we evaluated assumes full access to all benign gradients at the current step and robust aggregation strategy used in the server \textit{(agr-updates)}.

\section{Defenses Formulation} \label{sec:defenses}

A robust aggregator would take all the clients' update vectors and is expected to return an update $\widetilde{\textvec{g}}$ that is closer to true gradients and devoid of any potential corruptions.

\begin{equation}
\widetilde{\textvec{g}}^{(t)} = \textaggfunc{RAggr}( \{\textvec{g}^{(t)}_k : k \in \textset{H}\uplus\textset{B}\})
\label{eq:robustaggr-output}
\end{equation}
For the subsequent section, time step $t$ is skipped for convenience, and all the aggregation is at timestep $t$ gathering $t^{th}$ parameters of local clients unless indicated otherwise.

\subsection{Krum}
Krum\cite{defns-blanchard2017machine_krum} is a middle-seeking aggregator that selects a gradient that is closest to the most client vectors. The method computes the distance score ($\varsigma$) for each client based on its neighbors, and the client with a very minimal score is taken as a robust aggregate. 
In the Multi-Krum version, instead of taking a single client's update as aggregate, the average of top-$m$ clients with the minimum distance is returned as aggregate.

\begin{equation}
\begin{aligned}
     \widetilde{\textvec{w}}=&\; \textaggfunc{Krum} ( \{\textvec{w}_k : k \in \textset{K}\} ) := \textvec{w}_{i^\star} \\
           i^\star =&\;  \textstat{argmin}_{\substack{i \in \textset{K}}} (\varsigma_1,\ldots,\varsigma_{\textcount{K}}) \\
     \varsigma_i =&\; \sum_{i \to k} \|\textvec{w}_i - \textvec{w}_k \|^2  \qquad \forall i,k \in \textset{K}\\
\end{aligned}
\end{equation}

Notation ${i \to k}$ indicates the nearest neighbours for $i$ of count of which is $[\textcount{K} - \textcount{B} - 2]$.
The $i^\star$ is specific client for which condition $\varsigma_{i^\star} \leq \varsigma_{i}$ holds for all  $i \in \textset{K}$. Multi-Krum selects $m$ weight vectors  
$\{\textvec{w}^{\star}_1,\ldots,\textvec{w}^{\star}_m \}$ that has minimal distance scores and takes the average of these vectors $\frac{1}{m} \sum_{i=0}^m{\textvec{w}^{\star}_i}$ as aggregate. 
In our experiments, we take $m=1$.

\subsection{Coordinatewise Trimmed Mean (CWTM)}
CWTM \cite{defns-yin2018byzantine_cwtm} is a middle-seeking approach on individual vector elements across clients, respectively, instead of the whole vector. The median version selects single median values of gradients for each coordinate, discarding the rest of the gradients. For a set of client gradient vectors $\textvec{g}_k \in \mathbb{R}^\theta$, the coordinate-wise median is defined as follows: 

\begin{equation}
\begin{aligned}
    \widetilde{\textvec{g}} =&\; \textaggfunc{CwMed}(\{ \textvec{g}_k:k \in \textset{K} \}) := [\widetilde{\textvec{g}}^{0}, \widetilde{\textvec{g}}^{2},\dots,\widetilde{\textvec{g}}^{\textcount{D}}] \\
    \widetilde{\textvec{g}}^\varphi =&\; \textstat{median}( \{ \textvec{g}^\varphi_k : k \in \textset{K} \}) \quad \forall \varphi\: \in [\textcount{D}]
\end{aligned}
\end{equation}

For each coordinate $\varphi$ of $\widetilde{\textvec{g}}$ is calculated by taking the median of the corresponding coordinates from all vectors $\textvec{g}_k$, where $\textstat{median}$ represents the median operation on a given set of numbers.

The same formulation can be extended to trimmed mean statistics, where at each coordinate $\beta$ amount, gradients in both extremities are discarded, and the rest of the gradients are averaged.
This is more advantageous than the median as richer gradients from other clients can also contribute. The motivation is that Byzantine clients will have their gradients far away from the honest clients' gradients, \ie in the extremities.
Here each coordinates $\varphi$ is computed as the mean of $\{ \textvec{g}^{\varphi}_k; k \in \textset{U}^{\varphi}\}$ where $\textset{U}^{\varphi}$  is obtained by removing $\beta$ fraction of largest and smallest elements from $\{\textvec{g}^{\varphi}_1, \textvec{g}^{\varphi}_2, ...,\textvec{g}^{\varphi}_\textcount{K} \}$ for a given coordinate $\varphi$

\begin{equation}
\begin{aligned}
     \widetilde{\textvec{g}} =&\; \textaggfunc{CwTrmean} ( \{\textvec{g}_k : k \in \textset{K}\} ) :=\;[\widetilde{\textvec{g}}^{1},\ldots,\widetilde{\textvec{g}}^{\theta}] \\
     \widetilde{\textvec{g}}^{\varphi} =&\; \frac{1}{{\textcount{K}\cdot(1-2\beta)}} \sum_{ g \in \:\textset{U}^{\varphi}} g \;, \qquad \forall \varphi\: \in [\theta] \\
     \textset{U}^{\varphi} =& \; \{  \textvec{a}^{\varphi}_i:\; \beta\textcount{K} < i \leq (1-\beta)\textcount{K}\}\; \quad \forall \varphi \in [\theta] \\
      \textvec{a}^{\varphi} =& \; \textstat{sort}[\textvec{g}^{\varphi}_1, \textvec{g}^{\varphi}_2,\ldots,\textvec{g}^{\varphi}_\textcount{K} ] \\
\end{aligned}
\end{equation}

Here $\beta \in [0, \frac{1}{2})$ indicates the proportion of possible Byzantines in collaboration, thus malicious gradients can be discarded. For our experiments, we set $\beta = 0.2$.

\subsection{COPOD Distance-based Outlier Suppression (COPOD-DOS)}

This method\cite{defns-alkhunaizi2022suppressing_copod} detects and suppresses the outliers among the clients' parameters based on their pairwise distance. It uses both cosine similarity and 
Euclidean distance measures for identifying the outlier. This approach applies $\textstat{COPOD}$ outlier detection method on the pairwise distance matricies ($\textvec{M^S}, \textvec{M^E}$) and produces an outlier score $\textvec{r}=[r_1,\ldots,r_\textcount{K}]$ where $r_i \in (0,\infty)$ for each client. 
The outlier score is used to create weightage ($\lambda$) for each client that sums to 1 by applying \textstat{softmax} for aggregation.

\begin{equation}
\begin{gathered}
     \widetilde{\textvec{w}} =\; \textaggfunc{CopodDos} ( \{\textvec{w}_k : k \in \textset{K}\} ) :=  \sum_{k\in \textset{K}} \lambda_k \textvec{w}_k\\
     [\lambda_0,\ldots,\lambda_k]  =\; \textstat{softmax}(\textvec{r}) \qquad  \text{where}\quad \textvec{r} = \frac{\textvec{r^S} + \textvec{r^E}} {2}  \\
     \textvec{r^S} = \textstat{COPOD}(\textvec{M^S}) \quad;\quad 
     \textvec{M^S} = \bigl[ a_{ij}^\textvec{S}\bigr] \quad \\
     \textvec{r^E} = \textstat{COPOD}(\textvec{M^E}) \quad;\quad
     \textvec{M^E} = \bigl[ a_{ij}^\textvec{E}\bigr] \quad \\
     a_{ij}^\textvec{S} = 1 -\frac{\textvec{w}_i^\top \cdot \textvec{w}_j}{\|\textvec{w}_i\| \: \|\textvec{w}_j\|} \quad i,j \in [\textcount{K}] \\
     a_{ij}^\textvec{E} = \|\textvec{w}_i - \textvec{w}_j\| \quad i,j \in [\textcount{K}] \\
\end{gathered}
\end{equation}

\subsection{Robust Federated Aggregation - Geometric Median (RFA)}
Geometric Median is statically more robust than other methods leveraging the centrality of parameter distribution in clients. But the critical problem is that achieving this minimization problem computationally is tricker and hinders the practicality. RFA\cite{defns-pillutla2022rfa_geomed} method handles this as a minimization problem and utilizes Smoothed Weiszfeild's algorithm to approximate the geometric median $\textvec{v} $ in a faster and most stable manner.

\begin{equation}
\begin{aligned}
     \widetilde{\textvec{w}} =&\; \textaggfunc{GeoMedian} ( \{\textvec{w}_k : k \in \textset{K}\} ) \\
     :=&\;  \arg\!\min_{\textvec{v}} \sum_{k\in \textset{K}} \|\textvec{v} - \textvec{w}_k \|\\
     \textvec{v}^{(r+1)} =& \frac{\sum_{i=1}^\textcount{K} \beta_i^{(r)}\cdot \textvec{w}_i}{\sum_{i=1}^\textcount{K} \beta_i^{(r)}}, \quad\forall r \in [1,\ldots,\textcount{R}]\\
     \text{where} &\quad \beta^{(r)}_i = \frac{\alpha_i}{\textstat{max}[\varepsilon, \|\textvec{v}^{(r)} - \textvec{w}_i\|]} \quad\text{,}\quad \alpha_i = \frac{1}{\textcount{K}}
\end{aligned}
\end{equation}

The median is computed iteratively with $\textcount{R}$ steps. In our experiments, we set the Weiszfeld iterations $\textcount{R}=3$.

\subsection{Centered Clipping (CClipping)}

Centered Clipping \cite{defns-karimireddy2021learning_clipping} involves rescaling the gradient vector if it exceeds a predetermined radius from the reference vector and retaining it in the original scale if it is within the threshold. The underlying principle of this approach is that malicious gradients have to be far away from the reference vector ($\textvec{v}$), leading to excessive Euclidean distance, which will trigger clipping.
This method only scales back to a specific range and, thus does not exclude or suppress updates during aggregation from any clients, hence all diverse updates contribute to the aggregate vector.
Clipping is carried out iteratively with $\textcount{Q}$ steps during each aggreagation round. This further improves the convergence of training with enhanced stability and robustness while also ensuring convergence guarantees.
This work also explores momentum-based gradient updates from honest workers so that the variance of each global aggregation is bounded, hence further reducing the scope of the attack. For our evaluation, we reset the momentum in the optimizer during each global round to make the comparison fair with other approaches.
Global aggregation at $t$-th round can be given as

\begin{equation}
\begin{gathered}
     \widetilde{\textvec{g}}^{(t)} =\; \textaggfunc{CClip}_\textcount{Q} ( \{\textvec{g}^{(t)}_k : k \in \textset{K}\},\; \widetilde{\textvec{g}}^{(t+1)} ) := \textvec{v^{(\textcount{Q})}  }\\
     \textvec{v}^{(q)}  =\;
        \textvec{v}^{(q-1)} + \frac{1}{\textcount{K}} \sum_{k=1}^{\textcount{K}} 
        \left( (\textvec{g}^{(t)}_k - \textvec{v}^{(q-1)}) \cdot \varkappa  \right), \\
        \varkappa = \textstat{min}\Bigl[1, \frac{\tau}{\Vert\textvec{g}^{(t)}_k - \textvec{v}^{(q-1)} \Vert} \Bigr]  \quad \forall q \in [1,\ldots,\textcount{Q}]\\
\end{gathered}
\end{equation}

for the first clipping iteration gradient from the previous aggregation is used as reference estimate, $\textvec{v}^0 = \widetilde{\textvec{g}}^{(t+1)}$. We set clipping iterations as $\textcount{Q}=3$ and the clipping radius as $\tau=100$

\subsection{Randomized Bucketing with Centered Clipping (CC-RandBucket)}

Randomized bucketing \cite{defns-karimireddy2020byzantine_randBuck} approach splits the clients randomly into $\textcount{L}$ buckets, with each bucket holding $\textcount{S}$ client vectors. Each bucket is averaged into a single vector $\textvec{u}_l$ before applying aggregation methods such as $\textaggfunc{CClip}$. 
By randomly splitting clients into buckets
(a)~There is a higher chance of malicious updates grouped in different bins with the majority of regular clients; therefore reducing the influence of malicious clients would be reduced 
(b)~As the clients in buckets are averaged, at least a few of the clients' gradients are mixed and contribute to the aggregate vector even if $\textaggfunc{Aggr}$ follows median approaches.
This strategy also reduces the variance of gradients (among $\textvec{u}_l$) that are passed to the aggregator. Also, because of the mixing of representative gradients, over multiple rounds, the aggregate vector will be closer to the true gradient, especially in the absence of malicious updates and with mild heterogeneity.

\begin{equation}
\begin{aligned}
    \widetilde{\textvec{g}}^{(t)} =&\; \textaggfunc{CCRandB} ( \{\textvec{g}_k^{(t)} : k \in \textset{K}\} ) \\
    :=&\; \; \textaggfunc{CClip}_\textcount{1}( [\textvec{u}_1,\ldots,\textvec{u}_{\textcount{L}}] ,\;
    \widetilde{\textvec{g}}^{(t+1)} ) \\
    \textvec{u}_l =&\; \frac{1}{\textcount{S}} \sum_{i = \textcount{S}\cdot(l-1)+1}^{\textstat{min}[\textcount{K} ,\; \textcount{S}\cdot l] } 
    \textstat{shuffle}(\textvec{g}_1^{(t)},\ldots,\textvec{g}_\textcount{K}^{(t)})_i \\
    & \qquad \forall\; l \in [1,\ldots,\textcount{L}] ;\quad \textcount{L}= \lceil {\textcount{K}} / {\textcount{S}}\rceil \\
\end{aligned}
\end{equation}

For our experiments we set the clipping iterations $\textcount{Q}=1$, clipping radius $\tau=100$ and vectors per bucket $\textcount{S}=2$.

\subsection{Sequential Bucketing with Centered Clipping (CC-SeqBucket)}

Sequential Bucketing \cite{defns-ozfatura2023byzantines_seqbuck} takes a sequential aggregation procedure rather than a single shot approach like in  $\textaggfunc{CCRandB}$.
Total buckets is $\textcount{R}$ with each bucket holding $\textcount{S}$ client vectors.
When Byzantines count is substantially higher, owing to randomness they can form a part in the majority of the bucket affecting the aggregation. For example, in $\textaggfunc{CClip}$ knowledge starting reference point (which will be the previous round aggregate) can make the aggregation vulnerable to attack.
So $\textaggfunc{CCSeqB}$ provides additional guardrails to fix the above problems, 
(i)~Instead of random bucketing, uses similarity measures and partitions the vectors into buckets ($\textset{G}_r$) such that each bucket will include the most dissimilar vectors, therefore Byzantines cannot choose a unanimous direction and still form a majority within buckets.
(ii)~$\textaggfunc{Aggr}$ is applied to each bucket instead of simple averaging thus neutralizing deceitful updates at the bucket level.
(iii)~When clipping instead of having the same reference vector for each bucket, use aggregate from the previous bucket $\textvec{u}^{(r-1)}$ as a reference vector, and hence reference keeps shifting for each bucket, and its randomized, hence attacker cannot tailor anything specific to reference vector.

\begin{equation}
\begin{aligned}
    \widetilde{\textvec{g}} =&\; \textaggfunc{CCSeqB} ( \{\textvec{g}_k : k \in \textset{K}\} ) := \textvec{u}^{(\textcount{R})}\\
    \textvec{u}^{(r)} =&\; \textaggfunc{CClip}_\textcount{1}( \{ \textvec{g}: \forall \textvec{g}\in \textset{G}_r \},\; \textvec{u}^{r-1}) \\
    & \qquad \forall\; r \in [1,\ldots, \textcount{R} ]  
    \quad \textcount{R} = \lceil {\textcount{K}} / {\textcount{S}}\rceil \\
    \textset{G}_r =&\; \{\textvec{g}: 1\leq s \leq \textcount{S}, \: \textstat{if}\;\exists\; \textvec{g} \sim \textset{A}_s \} \\
    & \qquad \forall\; r \in [1,\ldots, \textcount{R} ]   \\
    \textset{A}_s =&\; \bigl\{\textstat{sort}_\textfunc{Y}
    (\textvec{g}_1,\dots,\textvec{g}_\textcount{K})_i:\; \\
    & {[\textcount{R}\cdot(s-1)+1] \leq i \leq \textstat{min}[\textcount{K},\: \textcount{R}\cdot s] }\bigr\} \\
    & \quad \forall\; s \in [1,\ldots,\textcount{S}]\\
    \text{where},&\quad \textfunc{Y}(\textvec{a}) \triangleq \frac{\textvec{a} \cdot \widetilde{\textvec{g}}^{(t+1)}}{\|\textvec{a}\| \|\widetilde{\textvec{g}}^{(t+1)}\|}\\
    \\
\end{aligned}
\end{equation}

Here, $\textfunc{Y}(\textvec{a})$ function is the score to sort the gradients.
We use clipping iterations $\textcount{Q}=1$, clipping radius $\tau=100$, vectors per bucket $\textcount{S}=2$.

\subsection{TIES-Merging with Robustness}

TIES-Merging \cite{merge-yadav2023resolving_ties} was introduced in the field of model merging as the solution to prevent performance drop due to information loss caused by interference of parameters. They identify deterioration mainly due to redundant parameter values and conflicting signs among the important parameters. The proposed solution does sign election based on the magnitude of the task vectors. The task vector is the difference between the initial model vector and the trained model vector, which is very much equivalent to gradients computed in a single update round. 
When adapting it as a defense against model poisoning, we cannot use the magnitude-based election approach since malicious clients can scale attack gradients, making it dominant. Thus, we modified the TIES-Merging not to use magnitude but instead to give equal weight to all gradients and make it conducive to FL, which we define as  $\textaggfunc{R-TiesMerge}$. We use this modified method for comparison with other $\textaggfunc{RAggr}$
In our proposed method, we compute the weighting explicitly using the Concordance Ratio, which differs from the TIES-Merging.

\begin{equation}
\begin{gathered}
    \widetilde{\textvec{g}} = \textaggfunc{R-TiesMerge}(\{\textvec{g}_k : k \in \textset{K}\}) 
    := [\widetilde{\textvec{g}}^{1},\ldots,\widetilde{\textvec{g}}^{\textcount{D}}] \\
    \widetilde{\textvec{g}}^\varphi = \frac{ \sum_{k=0}^{\textcount{K}} {\delta}^\varphi_k \cdot \textvec{v}_k^\varphi }{ \sum_{k=0}^{\textcount{K}} {\delta}^\varphi_k } 
    \qquad\text{where}\quad \delta^\varphi_k = \textit{I}\left(\textvec{u}^\varphi\cdot \textvec{v}_k^\varphi > 0 \right) \\
\end{gathered}
\label{eq:ties-merging}
\end{equation} 

Here $\delta \in \{0,1\}$ is an indicator variable defined to represent the alignment of signs. $\textvec{u}$ and $\textvec{v}$ are defined as follows, 

\begin{equation}
\begin{aligned}
    \textvec{u} &= [\xi^1, \ldots, \xi^\textcount{D}] = \textstat{sgn}\Bigl[\sum_{k} {\textstat{sgn}(\textvec{g}_k}) \Bigr] \qquad k \in \textset{K} \\
     \textvec{v}_k &= \textstat{sparsify}(\textvec{g}_k, \gamma_{\texttt{t}}) := [v^\varphi:\: \varphi = 1\ldots\textcount{D}]\\
    & \text{where}\;\; v^\varphi = \begin{cases}
                \textvec{g}^{\varphi}_k & \text{if } \textstat{abs}(\textvec{g}^{\varphi}_k) > \textstat{quantile}( \textvec{g}_k, \gamma_{\texttt{t}}) \\
                0 & \text{otherwise}
        \end{cases} \\
\end{aligned}
\label{eq:ties-sparse-elect}
\end{equation}  

Here $\gamma_{\texttt{t}}$ is the sparsification factor which is set at $0.9$, and vector $\textvec{u}$ represents the elect sign for each parameter in the model.
  
\subsection{Huber Loss-Based Weiszfeld Aggregation}
This method minimizes multi-dimensional Huber loss ($\Phi$) in the server aggregator. Actual implementation uses a modified Weizfield algorithm for the approximation indicated in Eq.\ref{eq:huberloss-approx}. It considers two requirements for good aggregator consistency without attack and robustness under attack through L2-loss minimization and geometric median.

\begin{equation}
\begin{aligned}
    \widetilde{\textvec{w}} =&\; \textaggfunc{HuberLossAgg} ( \{\textvec{w}_k : k \in \textset{K}\} ) \\
     :=&\;  \arg\!\min_{\textvec{v}} \sum_{k\in \textset{K}}  {\Phi}_{k}( \|\textvec{v} - \textvec{w}_k \|)\\
    & \text{where}\; {\Phi}_{k} = \begin{cases}
                \frac{1}{2} \textvec{u}^2 & \text{if } |\textvec{u}| \leq T_i \\
                T_k \textvec{u} - \frac{1}{2}T_{k}^{2}  & \text{if } |\textvec{u}| > T_k
                \end{cases} \\
\end{aligned}
\label{eq:huberloss}
\end{equation}

\begin{equation}
\begin{aligned}
    \textvec{c}^{(r+1)} =&\; \frac{\sum\limits_{k=1}^\textcount{K} \textstat{min} \left( 1, \frac{\tau}{\|\textvec{c}^{(r)} - \textvec{w}_k\|} \right)\cdot\textvec{w}_k}{\sum\limits_{k=1}^\textcount{K} \textstat{min} \left( 1, \frac{\tau}{\|\textvec{c}^{(r)} - \textvec{w}_k\|} \right)}
\end{aligned}
\label{eq:huberloss-approx}
\end{equation}

In this, we set the Huber loss threshold $\tau$ as $0.2$, and set the maximum number of iterations ($r$) for Weizfield as $100$.

\subsection{FL-Detector}

FL-Detector tries to explicitly detect malicious clients based on the history of previous updates. It uses GAP statistics on suspicion scores (shown in Eq.\ref{eq:fldetect-suscore}) to identify malicious clients.

\begin{equation}
\begin{gathered}
    \textvec{S}^{(t)} = \{\textvec{s}^{(t)}_1,\textvec{s}^{(t)}_2,\dots,\textvec{s}^{(t)}_n\} \quad \text{where} \\
    \textvec{s}^{(t)}_i = \frac{1}{\textcount{T}} \sum_{r=0}^{N-1} \textvec{d}^{(t-r)}_i \\
    \textvec{d}^{(t)}_i = \|\hat{\textvec{g}}^{(t)}_i - \textvec{g}^{(t)}_i\|_2 \\
    \hat{\mathbf{g}}^{(t)}_i = \mathbf{g}^{(t-1)}_i + \textvec{H}^{(t)}\cdot \textvec{g}^{(t)}_i \\
\end{gathered}
\label{eq:fldetect-suscore}
\end{equation}

Here, $\textvec{H}$ Hessian vector calculated using L-BFGS\cite{defns-zhang2022fldetector}.
$\textcount{T}$ is the window size that we set as 10. On suspicion score, we run the \textit{GAP} statistics to obtain the number of clusters. If the cluster count is more than 1, then we assume that there is the possibility of malicious updates.
If so, we then apply \textit{K-means} clustering with a cluster size of 2 and take the clients in the cluster with an overall minimum suspicion score.




\end{document}